\documentclass{article}

\usepackage{arxiv}

\usepackage[utf8]{inputenc} % allow utf-8 input
\usepackage[T1]{fontenc}    % use 8-bit T1 fonts
\usepackage{url}            % simple URL typesetting
\usepackage{booktabs}       % professional-quality tables
\usepackage{amsfonts}       % blackboard math symbols
\usepackage{nicefrac}       % compact symbols for 1/2, etc.
\usepackage{microtype}      % microtypography
\usepackage{lipsum}

\usepackage{cite}
\usepackage{amsmath,amssymb,amsfonts}
\usepackage{algorithmic}
\usepackage{graphicx}
\usepackage{textcomp}
\usepackage{bmpsize}
\usepackage{xcolor}
\usepackage{lipsum}
\usepackage[colorlinks=true,urlcolor=black]{hyperref}
\usepackage{mathtools}

\usepackage{multirow}
\usepackage[T1]{fontenc}
\usepackage{url}
\usepackage{booktabs}
\usepackage{amsfonts} 
\usepackage{nicefrac} 
\usepackage{microtype}
\usepackage{adjustbox}
\usepackage{todonotes}
\usepackage{amsmath}
\usepackage{cleveref}
\usepackage{xcolor}
\usepackage[linesnumbered,ruled,vlined]{algorithm2e}
\usepackage{algorithmic}

\SetCommentSty{mycommfont}

\usepackage{subfig}
\usepackage{xcolor}
\usepackage{lipsum}
\usepackage{comment}
\usepackage{tabularx}
\newcolumntype{Y}{>{\centering\arraybackslash}X}

\title{Sponge Examples: Energy-Latency Attacks on Neural Networks}

\let\oldnl\nl% Store \nl in \oldnl
\newcommand{\nonl}{\renewcommand{\nl}{\let\nl\oldnl}}% Remove line number for one line

\newcommand{\eg}{\textit{e.g\@.}}

\newcommand{\etal}{\textit{et~al\@.}}
\newcommand{\ie}{\textit{i.e\@.}}

\author{
  Ilia Shumailov \\
  University of Cambridge\\
  \texttt{ilia.shumailov@cl.cam.ac.uk} \\
   \And
  Yiren Zhao \\
  University of Cambridge\\
  \texttt{yiren.zhao@cl.cam.ac.uk} \\
  \And
  Daniel Bates \\
  University of Cambridge\\
  \texttt{daniel.bates@cl.cam.ac.uk} \\
  \And
  Nicolas Papernot \\
  University of Toronto and Vector Institute \\
  \texttt{nicolas.papernot@utoronto.ca} \\
  \And 
  Robert Mullins \\
  University of Cambridge\\
  \texttt{robert.mullins@cl.cam.ac.uk} \\
  \And
  Ross Anderson \\
  University of Cambridge\\
  \texttt{ross.anderson@cl.cam.ac.uk} \\
}

\begin{document}
\maketitle

\begin{abstract}
The high energy costs of neural network training and inference led to the use of acceleration hardware such as GPUs and TPUs. While such devices enable us to train large-scale neural networks in datacenters and deploy them on edge devices, their designers' focus so far is on average-case performance. In this work, we introduce a novel threat vector against neural networks whose energy consumption or decision latency are critical. We show how adversaries can exploit carefully-crafted \textbf{sponge examples}, which are inputs designed to maximise energy consumption and latency, to drive machine learning (ML) systems towards their worst-case performance. Sponge examples are, to our knowledge, the first 
%piece of work to provide 
denial-of-service attack against the ML components of such systems. 
% They can be used to drain battery-powered devices, and also to delay decisions where a network has critical real-time performance tasks, such as in perception for autonomous vehicles. 

We mount two variants of our sponge attack on a wide range of state-of-the-art neural network models, and find that language models are surprisingly vulnerable. Sponge examples frequently increase both latency and energy consumption of these models by a factor of $30\times$. Extensive experiments show that our new attack is effective across different hardware platforms (CPU, GPU and an ASIC simulator) on a wide range of different language tasks. On vision tasks, we show that sponge examples can be produced and a latency degradation observed, but the effect is less pronounced.
To demonstrate the effectiveness of sponge examples in the real world, we mount an attack against Microsoft Azure's translator and show an increase of response time from $1$ms to $6$s ($6000\times$). 
We conclude by proposing a defense strategy: shifting the analysis of energy consumption in hardware from an average-case to a worst-case perspective. 
\end{abstract}

% keywords can be removed

\section{Introduction}

%As Amdahl's law reduces the benefit that general-purpose multicore CPUs can realise from having more transistors, computer architects are increasingly developing  specialized hardware to accelerate specific tasks. 

The wide adoption of machine learning has led to serious study of its security vulnerabilities. Threat vectors such as adversarial examples~\cite{biggio2013evasion,szegedy2013intriguing}, data poisoning~\cite{nelson2008exploiting,jagielski2018manipulating}, membership inference~\cite{shokri2017membership,salem2018ml,choo2020label} and fault injection attacks~\cite{hong2019terminalbrain} have been extensively explored. 
%\todo{Add the discussion of latency consequences}
These attacks either target the \textit{confidentiality} or \textit{integrity} of machine learning systems~\cite{biggio2018wild,papernot2016towards}. So what about the third leg of the security triad: their \textit{availability}? 
In this paper, we introduce an attack that increases the power drawn by neural networks and the time they take to make decisions. An adversary may mount our attack on a datacenter providing ML-as-a-Service to cause disruption, \ie~denial-of-service~\cite{palmieri2015energy}. 
Increasing the energy consumption of edge devices such as smartphones can drain their batteries and make them unavailable~\cite{martin2004denial}. Perhaps even more seriously, an attack that slows down decisions can subvert safety-critical or mission-critical systems.

Our key observation is that different inputs of the same size can cause a deep neural network (DNN) to use very different amounts of time and energy: this \textit{energy-latency gap} is the vulnerability we exploit. The gap exists because of specific optimisations in hardware (\eg~leveraging input sparsity) and algorithms (\eg~a variable number of passes through the network for an input of the same size). 

Our attack can be even more effective against the growing number of systems that use GPUs or custom hardware. Machine learning in general, and neural networks in particular, command workloads heavy in matrix algebra. GPUs were fundamental to the AlexNet breakthrough in 2012~\cite{krizhevsky2012imagenet}; in response to increasing demand, Google introduced TPUs to facilitate inference -- and training -- in its datacenters~\cite{jouppi2017datacenter}, while Apple introduced the Neural Engine to make its smartphones more energy-efficient for on-device deep learning~\cite{apple2017}. Hardware engineers explicitly target the Operations per Watt (OPs/W) performance of DNN processing. But by increasing complexity, optimisations tend to increase the attack surface. There is ample precedent elsewhere in computer engineering for optimisations widening the gap between average-case and worst-case performance in ways that a capable attacker can exploit: a recent example is Spectre~\cite{kocher2019spectre}, which exploits hardware speculation to launch a powerful timing side-channel attack. Security engineers therefore need to pay close attention to worst-case performance. In this paper, we start this process for the optimisations used to speed up modern machine learning in both hardware and algorithms. 

\textbf{Sponge examples} are designed to soak up energy consumed by a given neural network, forcing the underlying hardware system running DNN inference towards its worst-case performance. We present two ways of generating sponge examples, one gradient-based and one using genetic algorithms. The gradient-based approach requires access to DNN model parameters, while the genetic algorithm only sends queries to the model and evolves inputs based on energy or latency measurements. These two attack methods cover both White-box and Black-box attacks, and our extensive experiments demonstrate the effectiveness of sponge examples under both scenarios. When considering modern ML-as-a-Service in a Black-box setting, our genetic algorithm successfully produces sponge examples that consistently increase the service response time and thus the energy consumption of the remote server. 

% Our first attack uses a genetic algorithm to craft malicious inputs, which we call \textit{\textbf{sponge examples}}, as they are designed to soak up energy from a neural network. Time or energy measurements obtained by profiling mutated inputs serve as the fitness function for the genetic algorithm to evolve better sponge examples. We find that this attack works against modern ML-as-a-Service in a black-box setting and can be used to perform an effective Denial-of-Service attack. 

% Our second attack optimizes sponge examples with L-BFGS and has an objective of encouraging inputs to have large activation norms across all hidden layers of a neural network. We find that this is even more effective at maximizing energy consumption during inference\footnote{Technically, we might describe this process as pessimization rather than optimization, as we're finding the inputs that give the worst possible performance.}. As illustrated in \Cref{fig:sponge_inforgraphics}, quite apart from battery-draining attacks, sponge examples could cause a control system to fail to meet real-time service requirements. For instance, autonomous vehicle designers should consider whether sponge examples could cause accidents, and engineers integrating DNNs into cognitive radar should analyse whether sponge examples create a new opportunity for jamming.

% \input{sections/images/infographics/sponge_diagram}

In this paper we make the following contributions: 
% Our contributions are the following: 
\begin{itemize}
    \item We introduce a novel threat against the availability of ML systems based on energy and latency. Our {\em sponge examples} are designed to cause inference to take as long as possible and consume as much energy as possible.
    
    \item We show that sponge examples cause increased energy consumption and longer run-time for a wide range of vision and language models. Sponge examples are particularly powerful against language models.
% can't say 30 times here after 6000 times in abstract! – was "increasing their energy consumption by up to $30\times$ and latency by up to $26\times$"
% \item  for both vision and language tasks are vulnerable to sponge examples, including compressed models specifically designed for edge devices (e.g., MobileNet). Our genetic algorithm attack increases the energy consumption of natural language tasks by up to $30\times$ and latency by up to $26\times$, while our enhanced L-BFGS attack increases the energy consumption of computer vision tasks by up to 8$\%$.
    
    \item We demonstrate the portability of sponge examples, by showing they are not only transferable across hardware platforms (CPUs, GPUs, and an ASIC simulator) but also across model architectures.
% \item showing that our conclusions are consistent across a variety of chips: CPUs, GPUs, and an ASIC simulator. Sponge examples also transfer across model architectures.

    \item We show that modern ML-as-a-Service is vulnerable to sponge attacks. With a 50-character input to Microsoft Azure translator, we can increase the latency from 1ms up to 6s: a $6000 \times$ degradation.
    
    \item We present a simple defense against sponge examples in the form of a worst-case performance bound for some models. This can also prevent unexpected increases in energy consumption in the absence of adversaries, potentially reducing the carbon footprint of models deployed for inference at scale.% (e.g., computer vision on smartphones and translation query services in the cloud). 
\end{itemize}

\section{Motivation}

Artificial Intelligence and Machine Learning have enabled real progress in automation, unleashing greater productivity and potential economic growth. Yet modern machine learning has become extremely power-hungry. It is estimated that the energy consumed in training a single transformer model is equivalent to 60\% of a car's lifetime carbon emissions~\cite{strubell2019energy}. And the energy cost doesn't stop there -- each inference consumes significant energy and happens ever more often\footnote{For example, OpenAI greatly limits number of queries one can do to the GPT3 model.}. Energy is the lifeblood of modern machine learning, as energy-efficient inference and training are needed to scale machine learning to more use cases. In this paper we explore the adversarial manipulation of energy and latency in ML models, services and applications. 

Modern hardware exploits many different optimisation techniques to maintain a high ratio of useful work to energy consumed. This often involves predicting future workloads and scheduling resources according to dynamic needs. Prediction and speculation occur at a number of levels in the stack, widening the gap between average-case and worst-case scenarios. This is particularly an issue in time or energy sensitive tasks, such as time series forecasting for automatic trading \cite{bernal2012financial} and activity recognition on wearable devices \cite{edel2016binarized}. In such applications, hitting worst-case performance could cause failures in decision making or deplete the batteries of user devices.
In safety-critical and real-time systems, such as autonomous vehicles which depend on scene understanding with tight latency constraints, service-denial attacks can pose a threat to life.
% Finally, the attacker can significantly affect usability of ML-based products. If a visual ad-blocker causes much longer to run users will stop using it.  

In this paper, we show that a capable attacker can exploit the performance dependency on hardware and model optimisations to launch a number of different attacks. We find that they can negate the effects of hardware optimisations, increase computation latency, increase hardware temperature and massively increase the amount of energy consumed. To make things worse, they can do this with very few assumptions, making attacks scalable in the real world. We further highlight the realism of sponge examples in a case study with Microsoft Azure translator, where we degraded latency up to a factor of $6000\times$.

On a number of occasions, despite the hardware protection provided by GPU engineers, we were able to increase temperature so that it passed the throttling point and sometimes even crashed the GPU drivers. The energy consumed by sponge examples on a machine learning model can therefore affect the underlying hardware if its power management software or hardware is not designed with adversaries in mind.

\section{Background}

\subsection{Hardware Acceleration for Deep Learning}
\label{sec:related:hardware}
Deep Neural Network (DNN) inference is both compute-intensive and memory-intensive.
Common hardware products such as CPUs and GPUs are now being adapted to this workload, and provide features for accelerating it.
Intel's Knights Mill CPU provides a set of SIMD instructions \cite{cooperation2016intel}, while NVIDIA's Volta GPU introduces Tensor Cores to facilitate the low-precision multiplications that underpin much of deep learning \cite{markidis2018nvidia}.

Hardware dedicated to deep learning is now pervasive in data centers, with examples including Big Basin at Facebook \cite{hazelwood2018applied}, BrainWave at Microsoft \cite{chung2018serving}, and racks of TPUs at Google \cite{jouppi2017datacenter,jouppi2018motivation}; the underlying hardware on these systems are either commodity hardware (Big Basin), re-configurable hardware (FPGAs for BrainWave), or custom silicon (TPUs). The latter two are specifically designed to improve the number of Operations per Watt of DNN inference. Careful modeling of average hardware efficiency allows ML-as-a-Service providers to price their services per query, rather then per energy used.
As we discuss later, custom and semi-custom hardware will typically exploit sparsity in data and the adequacy of low-precision computations for DNN inference, reducing both arithmetic complexity and the amount of DRAM traffic, to achieve significantly better power efficiency \cite{chen2019eyeriss,han2016eie,zhao2019automatic}. 
Our attack targets these optimisations among others.

% Performance per watt is an important indicator for the efficiency of cloud infrastructure \cite{barroso2005price}.
% Power oversubscription is a popular method for cloud services to handle provisioning.
% However, this makes data centers vulnerable
% to power attacks \cite{li2016power,somani2016ddos,xu2014power,xu2015measurement}.
% If malicious users can remotely generate power spikes on multiple hosts in the data center at the same time, they might overload the system and cause disruption of service~\cite{li2016power,palmieri2015energy}.
% Energy attacks against mobile devices aim to drain the battery more quickly~\cite{martin2004denial,fiore2014multimedia}.
% % Unlike energy attacks on large systems
% % increase the energy cost and occasionally cause denial of service,
% % if the victim is the mobile system,
% % battery depletion becomes a major concern.
% The possible victims of energy attacks on mobile systems range from phones to more constrained sensors~\cite{chen2009sensor}.
% Higher energy consumption also increases hardware temperature, which in turn increases the failure rate.
% For example, Anderson \etal~note that an increase of $15^\circ$C causes component failure rates to double~\cite{anderson2003more}.
% Modern hardware throttles to avoid over-heating; but throttling causes tasks to consume more energy as now they take longer to run, increasing total static and dynamic power costs.

% As neural networks becoming a major workload in both
% edge devices and the cloud, in this paper, from the security point of view,
% we show the energy characteristics of this new workload.

\subsection{Attacks on Energy}
\label{sec:related:energy}
%Our work links tightly to
%existing energy attacks on mobile devices
% and data centers in the field of security.

Operations per Watt are an important indicator of the efficiency of cloud infrastructure \cite{barroso2005price}.
Power oversubscription is a popular method for cloud services to handle provisioning, but it leaves datacenters vulnerable to power attacks \cite{li2016power,somani2016ddos,xu2014power,xu2015measurement}.
If malicious users can remotely generate power spikes on multiple hosts in the data center at the same time, they might overload the system and cause disruption of service~\cite{li2016power,palmieri2015energy}.
Energy attacks against mobile devices usually aim to drain the battery more quickly~\cite{martin2004denial,fiore2014multimedia}, although energy management in mobile devices can also be used to perform deterministic fault injection~\cite{tang2017clkscrew}.
%Unlike energy attacks on large systems increase the energy cost and occasionally cause denial of service, if the victim is the mobile system, battery depletion becomes a major concern.
The possible victims of energy attacks on mobile systems range from autonomous vehicles to sensors with constrained computing abilities \cite{chen2009sensor}.
Higher energy consumption also increases hardware temperature, which in turn increases the failure rate.
For example, Anderson \etal~note that an increase of $15^\circ$C causes component failure rates to go up by $2 \times$\cite{anderson2003more}.
Modern hardware throttles to avoid overheating; while short-term power savings may be possible through such voltage scaling, the overall energy consumption increases~\cite{efraim2014energyaware}. This creates nonlinear dependencies between energy and latency.
% As neural networks becoming a major workload in both
% edge devices and the cloud, in this paper, from the security point of view,
% we show the energy characteristics of this new workload.

\subsection{Security of Machine Learning}

% \todo{lets make a reference to zip bombs here}
Machine learning has been shown to be vulnerable to a number of different attack vectors~\cite{tabassi2019taxonomy}. Adversarial examples can cause a system to classify inputs incorrectly~\cite{biggio2013evasion,szegedy2013intriguing}. Adversarial examples can be found in the White-box setting through gradient-based optimization~\cite{biggio2013evasion,szegedy2013intriguing} while in the Black-box setting, the adversary can transfer adversarial examples from another model~\cite{papernot2017practical} or approximate gradients with finite differences~\cite{chen2017zoo} when they can observe the model's confidence as well as the output label. Data can be poisoned to manipulate future  performance~\cite{nelson2008exploiting,jagielski2018manipulating}. Run-time bit-errors can be introduced to greatly reduce  performance~\cite{hong2019terminalbrain}. These attacks either target the \textit{confidentiality} or \textit{integrity} of machine learning systems~\cite{biggio2018wild,papernot2016towards}. 

Here, we explore availability, \ie~timely and reliable access to information~\cite{nieles2017introduction}, and introduce a new form of service denial with samples that act as a sponge for time or energy. Service-denial attacks are well known in the context of computer networking~\cite{ferguson2000rfc2827,bellardo2003802}, but have been overlooked so far in ML. The current NIST draft on adversarial machine learning touches upon availability, but does not provide any examples of attacks~\cite{tabassi2019taxonomy}. 

Poisoning can perhaps be seen as an availability attack. If an attacker can poison data so that the machine learning model stops training or does so with reduced accuracy, this may be seen in some contexts as reducing availability. For example, Erba et al. presented such an attack against Industrial Control Systems~\cite{erba2019realtimeevasion}. However, the attacks presented in this paper do not poison data, but target either the hardware or the algorithmic complexity of the model. 

\section{Methodology}
\label{sec:method}

\subsection{Threat Model}
\label{sec:adversary}

In this paper we assume an adversary with the ability to supply an input sample to a target system, which then processes the sample using a single CPU, GPU or ASIC.
We assume no rate limiting, apart from on-device dynamic power control or thermal throttling.\footnote{Thermal throttling refers here to the deliberate slow-down of device performance when cooling is no longer able to dissipate the heat generated by a workload.} We assume no physical access to the systems \ie~ an attacker cannot reprogram the hardware or change the configuration.

We consider three threat models.
The first is a \textbf{White-box} setup: we assume the attackers know the model architecture and parameters.
The second considers an \textbf{interactive Black-box} threat: we assume attackers have no knowledge of the architecture and parameters, but are able to query the target as many times as they want and to time operations or measure energy consumption remotely.
The third is the \textbf{blind adversary}: we assume no knowledge of the target architecture and parameters, and assume no ability to take direct measurements. In this setting, the adversary has to transfer previously-discovered sponge examples directly to a new target -- without prior interaction.

Our adversary models loosely capture ML-as-a-Service deployments and on-device data processing.
A simple example could be a dialogue or a translation system. Users interact continuously by sending queries and can measure energy consumption, or when that is not possible by the response time (see Section~\ref{sec:eval_nlp}). Indeed, in \Cref{sec:azure} we show on an example of a Microsoft Azure translator that modern ML-as-a-Service is vulnerable to sponge attacks which only rely on the adversary's ability to observe response latency---even in presence of networking delay. 

\subsection{The Energy Gap}

% The same sample can take a drastically different amount of energy to compute. The reasoning lies in the relationship between the powers. Both of them are dependant on the voltage, but static further depends on temperature (see equation~\ref{eqn:static}) and dynamic on the frequency(see equation~\ref{eqn:dynamic})~\cite{mair2014myths}. Temperature dependency is also exponential in nature and was previously shown to be significant in overall power consumption. For example, Kocanda and Kos reported from $\times2$ to $\times85$ power consumption increase for the same computation for temperature rising from $20^\circ C$ to $100^\circ C$~\cite{kocanda2015static}.

% Modern hardware employs
% a large number of optimisations to
% minimise the amount of energy consumed,
% including dynamic voltage scaling,
% clock gating and using multiple clock domains.
% All of these factors can greatly affect
% the power and latency of the workload.

The \textit{Energy Gap} is the performance gap between average-case and worst-case performance, and is the target for our sponge attacks. To better understand the cause of this  gap, we tested three hardware platforms: a CPU, a GPU and an ASIC simulator. The amount of energy consumed by one inference pass (\ie~a forward pass in a neural network) depends primarily on~\cite{horowitz2014computingenergy}: 
\begin{itemize}
    \item the overall number of arithmetic operations required to process the inputs; and
    \item the number of memory accesses \eg~to the GPU DRAM.
\end{itemize}

The intriguing question now is:
\begin{center}
\textit{is there a significant gap in energy consumption for different model inputs of the same dimension?}
\end{center}

As well as fixing the dimension of inputs, \ie~not increasing the number of characters in a text sample or the pixel dimension of an image, we also do not consider inputs that would exceed the pre-defined numerical range of each input dimension. If models do have a large energy gap between different inputs, we describe two hypotheses that we think attackers can exploit to create \textit{sponge examples}, that is, inputs trigger the worst-case performance and have abnormally high energy consumption. 

\subsubsection{Hypothesis 1: Computation Dimensions}
\label{sec:hypo1}

% In many cases, variations in data shapes
% result in a huge difference in all subsequent
% computation dimensions and algorithmic complexity.
% Rarely is energy directly considered
% when designing models
% rather through a proxy of latency.
% For example,
% criticism of recurrent models
% for NLP tasks was their inability to be executed in parallel,
% because computation is factored
% along the symbol positions
% in both input and output sequences.
% This latency issue was solved through the use of Transformers,
% eschewing recurrence and using solely attention,
% to achieve better model-level parallelism \cite{vaswani2017attention}.
% Transformer has both improved the task performance
% and energy consumption through a reduction of the execution time $t$.
% Yet, in this paper, we argue that these changes
% are not enough and instead of focusing on the average energy cost,
% worst case scenario might be utilised by attackers.

Aside from data sparsity, modern neural networks also have a computational dimension. Along with variable input and output shapes, the internal representation size often changes as well -- for example, in the Transformer-based architectures for machine translation~\cite{vaswani2017attention}. The model is autoregressive in this case; both the input and output are sequences of words and internal computation depends on both of them. Before text gets to the model it has to go through a number of stages. First, individual components are separated within the sentence, removing punctuation and keeping useful words. Next, each word is represented as a number of tokens whose shape depends on the richness of input and output dictionaries. Because we cannot represent words mathematically, we need to map them to some numerical form. Yet we cannot build a mapping with all possible words, because that greatly increases model complexity, so in practice dictionaries with most-popular sub-words are used. Once tokenized, individual tokens are then projected into the embedding space (\eg~word2vec~\cite{mikolov2013distributed}), a high-dimensional space where knowledge about individual tokens is encoded. As computation progresses, each inference step depends on the embeddings of all of input tokens and output tokens produced so far. For example, imagine encoding the word `Athazagoraphobia`. With commonly used English dictionaries, it will get assigned 4 tokens for its input size of 16: `ath`, `az`, `agor`, `aphobia`. If a user makes a typing mistake, say `Athazagoraph\textbf{p}bia`, then suddenly its representation turns into 7 tokens for the same size of 16: `ath`, `az`, `agor`, `aph`, `p`, `bi`, `a`. An adversary can exploit it and construct large token representations. For example, `A/h/z/g/r/p/p/i/` will be 16 separate tokens. Ultimately, unknown words both in the input and output spaces will lead to a much larger sentence representation and many more inference runs.

% we explain where additional energy costs can be acquired.
% Although the principals used to attack the translation tasks
% are not universal across all NLP tasks,
% we found a range of models that the attack can successfully generalise to.

Consider an input sequence $x$ and an output sequence $y$. We denote the input and output token sizes (\ie~ the number of individual tokens extracted from an input sentence and produced for the output sentence) with $l_{\mathsf{tin}}$ and $l_{\mathsf{tout}}$. Each of the words in a sequence is embedded in a space of  dimensionality $l_{\mathsf{ein}}$, for the input, and $l_{\mathsf{eout}}$, for the output. Algorithm \ref{alg:transformer} contains the pseudocode for a Transformer's principal steps. In red, we annotate the computational complexity of the following instruction. As can be seen, several quantities can be manipulated by an adversary to increase the algorithm's run time:
1) token size of the input sentence $l_{\mathsf{tin}}$;
2) token size of the output sentence $l_{\mathsf{tout}}$;
and 3) size of the input and output embedding spaces ($l_{\mathsf{ein}}$ and $l_{\mathsf{eout}}$).
All of the above can cause a non-linear increase in algorithmic complexity and thus heavily increase the amount of energy consumed.
Note that perturbing these quantities does not require that the adversary modify the dimension of input sequence $x$; that is, with no changes to the input length, the adversary can increase energy consumption non-linearly.

\SetKwInput{KwInput}{Input}

\begin{algorithm}
\SetAlgoLined
\KwInput{Text sentence x}
\KwResult{y}
\nonl \textcolor{red}{$\downarrow$ O($l_{\mathsf{tin}}$)}\\
$x_{\mathsf{tin}}$ = \textsf{Tokenize}(x);\\
$y_{\mathsf{touts}} = \emptyset;$ \\
\nonl \textcolor{red}{$\downarrow$ O($l_{\mathsf{ein}}$)} \\
$x_{\mathsf{ein}}$ = \textsf{Encode} ($x_{\mathsf{tin}}$); \\
\nonl \textcolor{red}{$\downarrow$ O($ l_{\mathsf{tin}} \times l_{\mathsf{ein}} \times l_{\mathsf{tout}} \times l_{\mathsf{eout}}$)}\\
\While{ $y_{\mathsf{tout}}$ has no end of sentence token}{
\nonl \textcolor{red}{$\downarrow$ O($l_{\mathsf{eout}}$)}\\
 $y_{\mathsf{eout}}$ = \textsf{Encode} ($y_{\mathsf{tout}}$);\\
\nonl \textcolor{red}{$\downarrow$ O($l_{\mathsf{ein}} \times l_{\mathsf{eout}}$)}\\
 $y_{\mathsf{eout}}$ = \textsf{model.Inference}($x_{\mathsf{ein}}$, $y_{\mathsf{eout}}$, $y_{\mathsf{touts}}$); \\
\nonl \textcolor{red}{$\downarrow$ O($l_{\mathsf{eout}}$)};\\
 $y_{\mathsf{tout}}$ = \textsf{Decode}($y_{\mathsf{eout}}$);\\
 $y_{\mathsf{touts}}$.add($y_{\mathsf{tout}}$);
}
\nonl \textcolor{red}{$\downarrow$ O($l_{\mathsf{tout}}$)};\\
y = \textsf{Detokenize}($y_{\mathsf{touts}}$)
\caption{Translation Transformer NLP pipeline}
\label{alg:transformer}
\end{algorithm}

% Ultimately, the best way to think about such attack is that longer sentences consume more, both as inputs and outputs. Further, the representation of those sentences should be considered as well.

% Explain what is an energy gap and what are possible causes of the energy gap in the algorithm level.
% \begin{itemize}
% \item Tokenisation in NLP models.
% \item Regular and irregular sparsities in model weights and activations.
% \item Masking operations
% \end{itemize}
% (Insert images with transformer and CNN)

% . It is no surprise that static power dissipation, a result of existing leakage currents, rises with temperature. When temperature rises from 20 to 100°C static power multiplicities at least 2 times up to 85 times~\cite{kocanda2015static}.

% The power consumption of GPUs can be divided into two parts, namely, leakage
% power and dynamic power. The dynamic power is a function of operating temperature
% and circuit technology. Leakage power is consumed when the GPU is powered, even if
% there are no runtime activities. The dynamic power arises from switching of transistors
% and is determined by the runtime activities. Different components such as SMs and
% memories (e.g., local, global, shared, etc.) contribute to this power consumption.

\subsubsection{Hypothesis 2: Data Sparsity}
\label{sec:hypo2}
The rectified linear unit (ReLU), which computes $x\mapsto \max(0, x)$, is the de facto choice of activation function in neural network architectures. This design introduces sparsity in the activations of hidden layers when the weighted sum of inputs to a neuron is negative.
A large number of ASIC neural network accelerators consequently exploit runtime data sparsity to increase efficiency \cite{sharma2018bit,parashar2017scnn,nikolic2019characterizing}. For instance, ASIC accelerators may employ zero-skipping multiplications or encode DRAM traffic to reduce the off-chip bandwidth requirement.
The latest Xilinx AI compiler provides optimisations \cite{kathail2020xilinx} for automatically deploying sparse models to their FPGA devices, promoting the use of model sparsity in production systems.
On the algorithmic level, there is a recent surge of interest in using dynamic coarse-grained sparsity for accelerating GPU inference \cite{gao2018dynamic,hua2019channel}.
Hence, inputs that lead to less sparse activations will increase the number of operations and the number of memory accesses, and thus energy consumption.

\subsection{The Laws of Physics}
\label{sec:methodology:definitions}
Before getting to the details of the attack, we need to understand what affects energy consumption and what the attacker can reliably influence. Energy $E$ is the total consumed static power $P_{\mathsf{static}}$ and dynamic power $P_{\mathsf{dynamic}}$ for an interval of time $t$.
This energy formulation can be analysed in more detail; we show how to do this in \Cref{apdx:sec:energy}. 

\begin{equation}
\begin{aligned}
\label{alg:energy}
E = & (P_{\mathsf{static}} + P_{\mathsf{dynamic}}) \times t \\
= & \big{(}[\sum{I_{s} \times (\overbrace{e^{\frac{qV_d}{kT}-1}}}^{\mathclap{\substack{%
\text{\color{blue}overheat or increase overall consumption}\\~\\~
}}})\times V_{\mathsf{core}}] \\ 
& + [\underbrace{\textstyle \alpha}_{\mathclap{\text{\color{blue} more activity of the board}}} \times C \times V^{2}_{\mathsf{core}} \times \overbrace{f}^{\mathclap{\text{\color{blue}throttle or exploit load predictor}}}]\big{)} \times \underbrace{t}_{\mathclap{\text{\color{blue}run for longer or exploit the predictor}}}.
\end{aligned}
\end{equation}

The salient elements from \Cref{alg:energy} are that an attacker can affect energy use through four parameters: $T$ (temperature), $\alpha$ (activity ratio), $f$ (frequency) and $t$ (time).
Our sponge examples directly exploit the activity ratio $\alpha$ and execution time $t$, since these two parameters are tightly linked to the number of operations and memory accesses performed by model inference.
Although frequency $f$ and temperature $T$ will be influenced indirectly through optimisations performed by the underlying hardware, these are not our direct targets.
We hypothesise these parameters ($f$ and $T$) can also be exploited to create hardware level availability attacks on ML systems, \eg ~forced throttling or heating of devices, but they are beyond the scope of this paper. 

\subsection{Attack Methods and Setups}
\label{sec:eee_algo}
Having presented the intuition behind our attacks, we now introduce strategies for finding sponge examples corresponding to the threat models described in~\Cref{sec:adversary}.

% The first, a white-box attack, uses knowledge of the model's architecture and parameters to maximize the model's energy consumption.
% The second operates in the Black-box threat model described in Section~\ref{sec:adversary}, where the adversary can measure the time or energy consumed during inference. 
% Finally, our third strategy demonstrates the transferability of sponges across both network architectures and hardware accelerators. It operates in the threat model of a blind adversary.

\subsubsection{Genetic Algorithms in White-box and Black-box Settings}
\label{sec:method:ga}
\begin{figure*}[h]
    \centering
    \includegraphics[width=0.8\linewidth]{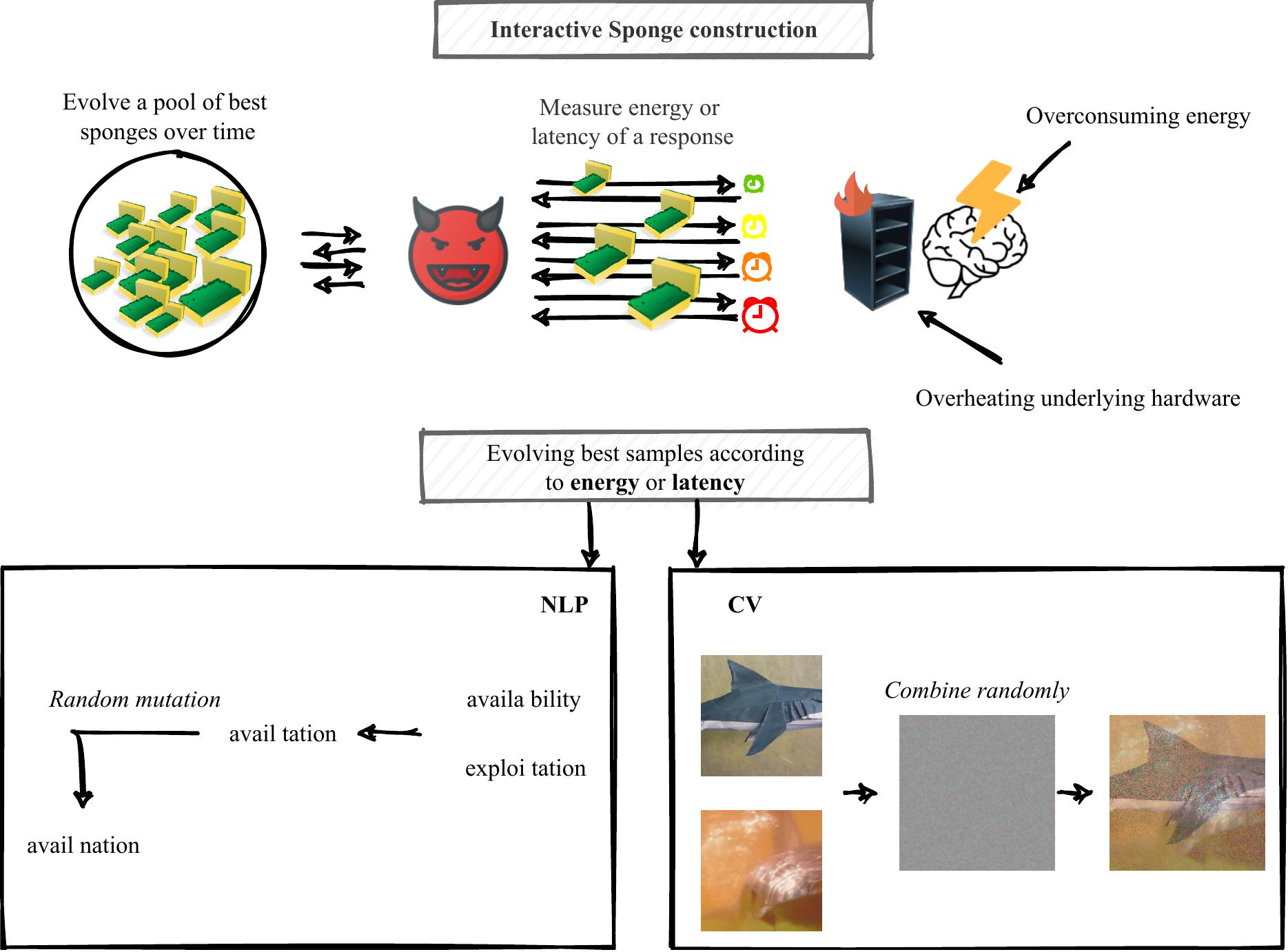}
    \caption{Availability adversary constructs sponge examples using a genetic algorithm. The adversary tries samples against the model and measures either latency or energy consumed, mixing the best performing samples in the pool. Eventually, the attacker identifies potent sponge examples.}
    \label{fig:interactive_infographics}
\end{figure*}

Genetic algorithms (GA) are a powerful tool for adversaries~\cite{xu2016automatically}. They can optimise a diverse set of objectives, and require no local gradient information. They are a particularly good fit for adversaries who only have access to the model's prediction in a Black-box setting. The general pipeline of a GA is presented in Algorithm~\ref{alg:ga} in Appendix. We start with a pool of randomly generated samples $S$. These are images for computer vision models, or sentences for NLP tasks. We then iteratively evolve the population pool as is depicted in~\Cref{fig:interactive_infographics}.
\begin{itemize}
    \item For computer vision tasks,
we sample two parents $A$ and $B$ from the population pool, and crossover the inputs using a random mask $A * \mathsf{mask} + (1-\mathsf{mask}) * B$. 
    \item For NLP tasks, we sample two parents $A$ and $B$, and crossover  by concatenating the left part of parent $A$ with the right part of parent $B$. We then probabilistically invert the two parts.
\end{itemize}
We explain the reasons for these choices in~\Cref{apdx:sec:reasons}. Next, we randomly mutate (\ie~ perturb) a proportion of the input features (\ie~ pixels in vision, words in NLP) of the children.
To maintain enough diversity in the pool, where applicable we preserve the best per-class samples in the pool. We obtain a fitness score $P$ for all pool members, namely their energy consumption.
We then select the winning top 10\% of samples $\hat{S}$,\footnote{As the sample pool is large, selecting the top 10\% makes the process more tractable.}, and use them as parents for the next iteration.
This genetic algorithm is simple but effective in finding sponge examples. Parameter choice is explained in~\Cref{apdx:sec:parameter_choices}.
In \Cref{apdx:sec:reasons}, we further explain the domain-specific optimisations of the GA algorithm on NLP and CV tasks for achieving a better attack performance.

Although following the same algorithm described above, we form two variants of GA for Black-box and White-box attacks respectively, each differing the way we measure fitness:
\begin{itemize}
    \item White-box GA: We access the parameters of the neural networks and provide an estimated energy cost based on the run-time sparsity,  \ie~number of operations based on the structure and parameters of the neural networks.
    \item Black-box GA: We do not access any of the neural network internals, and use purely the measured hardware cost as the fitness, \ie~latency or energy consumption. 
\end{itemize}

\subsubsection{L-BFGS in the White-box Setting}

We now consider an adversary with access to the model's parameters. Rather than a genetic algorithm, we use L-BFGS \cite{byrd1995limited} to optimise the following objective:
\begin{equation}\label{eq:lbfgsbloss}
    - \sum_{a_l \in A}{\|a_l\|}_2
\end{equation}
where $A$ is the set of all activation values and $a_l$ the activations of layer $l$. 
This generates inputs that increase activation values of the model across all of the layers simultaneously. Following Objective 1 outlined above, the increase in density prevents hardware from skipping some of the operations, which in turn increases energy consumption. We only evaluate the performance of sponge examples found by L-BFGS on computer vision tasks because of the discrete nature of the NLP tasks, which prevents differentiating the objective in~\Cref{eq:lbfgsbloss}\footnote{It is worth noting that for NLP tasks given knowledge of the dictionary an attacker can design the worst possible input and output token sequences. In this paper we make no assumptions about the dictionary or the model deployed, instead we optimise directly over energy or time.}.

\subsubsection{Cross-model and Cross-hardware Transferability for Blind Adversaries}

When adversaries are unable to query the model, they cannot
directly solve an optimisation problem to find sponge examples, even using the interactive Black-box approach, \ie~the GA. In this blind-adversary setting, we exploit transferability across both models \textit{and} hardware.
Indeed, in \Cref{sec:blackbox_hw_transfer} and \Cref{apdx:sec:cv_attack_transfer} in the Appendix, we show that sponge examples transfer across models. 
We examine three hardware platforms in our evaluation:
\begin{itemize}
    \item \textbf{CPU:} The platform is an Intel(R) Xeon(R) CPU E5-2620 v4 with 2.10GHz clock frequency. We use the Running Average Power Limit (RAPL) to measure energy consumption of the CPU. RAPL has been thoroughly evaluated and found to reflect actual energy consumption, as long as the counters are not sampled too quickly~\cite{hahnel2012measuring,khan2018raplinaction}.
    \item \textbf{GPU:} We use a GeForce 1080 Ti GPU with a 250.0 Watts power limit, a $96^\circ C$ slowdown temperature and a $84^\circ C$ throttling temperature. We use the NVIDIA Management Library (NVML) to measure energy consumption. NVML was previously found to capture energy quite accurately, with occasional instability for high-low patterns and high sampling rates~\cite{sen2018gpumeasure}.
    \item \textbf{ASIC:} We also developed a deterministic ASIC simulator, which monitors and records the runtime operations and number of DRAM accesses assuming a conservative memory flushing strategy. We then use measurements by Horowitz to approximate energy consumption~\cite{horowitz2014computingenergy}: at 45nm technology and 0.9V, we assume 1950 pJ to access a 32 bit value in DRAM and 3.7 pJ for a floating-point multiplication.
\end{itemize}
We show in~\Cref{sec:blackbox_hw_transfer} that sponge examples transfer across these types of hardware.

\section{Sponge Examples on Language Models}
\label{sec:eval_nlp}
\subsection{Models and Datasets}
We first evaluate our sponge example attack on a range of NLP models provided by the FairSeq framework~\cite{ott2019fairseq}. The models we consider have achieved top performance at their respective tasks and are used heavily in the real world to analyse data at scale. We report the performance of the RoBERTa~\cite{liu2019roberta} model, an optimised BERT~\cite{devlin2018bert}, on three GLUE benchmarks designed to assess language understanding~\cite{wang2018glue}. The datasets we considered include tasks in the SuperGLUE benchmark plus a number of machine-translation tasks.
The SuperGLUE benchmark follows the style of GLUE but includes a wider range of language-understanding tasks including question answering and conference resolution \cite{wang2018glue,wang2019superglue}. 
Further, we evaluate the attack on a number of translation tasks (WMT) using Transformer-based models~\cite{ott2018scaling,edunov2018understanding,ng2019WMT}. Both translation and language comprehension are fundamental to human society and form a bridge between computers and humans. They are built into virtual assistants and many other applications in the real world that are used on a day-to-day basis. 

Consider the pipeline for handling text. Before getting to the models, the text goes through several preprocessing steps. First, words get tokenized in a manner meaningful for the language. We used the tokenizer from the Moses toolkit\cite{koehn2007moses}, which separates punctuation from words and normalises characters. Next, tokenized blocks get encoded. Until recently, unknown words were simply replaced with an unknown token. Modern encoders improve performance by exploiting the idea that many words are a combination of other words. BPE is a popular approach that breaks unknown words into subwords it knows and uses those as individual tokens~\cite{sennrich2015subword}. 
In that way, known sentences get encoded very efficiently, mapping every word to a single token, and the number of computations is greatly reduced. 

\subsection{White-box Sponge Examples}
\label{sec:eval:language}

% In Section~\ref{sec:exp-setup} we described how words are broken into subwords before being encoded, which has the benefit of allowing a model to process unknown words but leads to larger sentence representations. Following the hypothesis from \Cref{sec:hypo2}, we demonstrate how this allows an attacker to mount our sponge samples to increase a model's energy consumption at a fixed input size. 

% \input{sections/tables/nlps_partial}

\begin{table*}[!h]
\centering
\begin{adjustbox}{scale=1.1,center}
% \begin{tabular}{@{}lcccccccccccc@{}}
\begin{tabularx}{\textwidth}{lcc *{9}{Y} }
\toprule
& 
& 
& \multicolumn{3}{c}{\textbf{GPU Energy [mJ]}}
& \multicolumn{3}{c}{\textbf{ASIC Energy [mJ]}}
& \multicolumn{3}{c}{\textbf{GPU Time [mS]}}
% & GPU 
\\
& Input size 
&
& Natural
& Random
& Sponge
& Natural
& Random
& Sponge 
& Natural
& Random
& Sponge 
% & Energy$\times$/Time$\times$
\\
% & $\text{Sponge}_{\text{GPU}}$ & $\text{Natural}_{\text{ASIC}}$ [mJ] & $\text{Random}_{\text{ASIC}}$ [mJ] & & $\text{Sponge Mean}_{\text{ASIC}}$ [mJ] & $\text{Sponge Top 10\%}_{\text{ASIC}}$ [mJ] & & 
%                 $\text{Energy}_{\text{ASIC}}$ &
%                 $\text{Energy}_{\text{GPU}}$ & & $\text{Time}_{\text{GPU}}$ \\
\midrule
\multicolumn{12}{l}{\textit{\underline{SuperGLUE Benchmark with \cite{liu2019roberta}}}} \vspace{3mm}\\
\multirow{6}{*}{CoLA}
& \multirow{2}{*}{15} &
& 2865.68   & 3023.705  & \multicolumn{1}{c|}{3170.38} 
& 504.93    & 566.58    & \multicolumn{1}{c|}{583.56}
& 0.02      & 0.02      & \multicolumn{1}{c}{$0.02$} \\
& & & 
$1.00\times$ & $1.06\times$     & \multicolumn{1}{c|}{$\mathbf{1.11}\times$} 
& $1.00\times$ & $1.12\times$   & \multicolumn{1}{c|}{$\mathbf{1.16}\times$}
& $\mathbf{1.00}\times$ & $0.92\times$   & \multicolumn{1}{c}{$0.92\times$}
% & $1.21$ 
\\
& \multirow{2}{*}{30} &
& 3299.07   & 4204.121  & \multicolumn{1}{c|}{4228.22} 
& 508.73    & 634.24    & \multicolumn{1}{c|}{669.20}
& 0.03      & 0.03      & \multicolumn{1}{c}{$0.02$} \\
& & 
& $1.00\times$  & $1.27\times$  & \multicolumn{1}{c|}{$\mathbf{1.28}\times$} 
& $1.00\times$  & $1.25\times$  & \multicolumn{1}{c|}{$\mathbf{1.32}\times$}
& $\mathbf{1.00}\times$ & $0.93\times$   & \multicolumn{1}{c}{$0.82\times$} 
% & $1.56$
\\
& \multirow{2}{*}{50} &
& 3384.62       & 6310.504      & \multicolumn{1}{c|}{6988.57}
& 511.43        & 724.48        & \multicolumn{1}{c|}{780.57}
& 0.03          & 0.04          & \multicolumn{1}{c}{$0.04$} \\
& & 
& $1.00\times$  & $1.86\times$  & \multicolumn{1}{c|}{$\mathbf{2.06}\times$} 
& $1.00\times$  & $1.42\times$  & \multicolumn{1}{c|}{$\mathbf{1.53}\times$} 
& $1.00\times$ & $1.23\times$ & \multicolumn{1}{c}{$\mathbf{1.27}\times$} 
% & $1.62$
\\ \\

\multirow{6}{*}{MNLI}
& \multirow{2}{*}{15} & 
& 3203.01       & 3573.93       & \multicolumn{1}{c|}{3597.3} 
& 509.19        & 570.10        & \multicolumn{1}{c|}{586.43} 
& 0.03          & 0.03          & \multicolumn{1}{c}{$0.03$} \\
& & 
& $1.00\times$  & $1.12\times$  & \multicolumn{1}{c|}{$\mathbf{1.12}\times$} 
& $1.00\times$  & $1.12\times$  & \multicolumn{1}{c|}{$\mathbf{1.15}\times$} 
& $1.00\times$  & $\mathbf{1.01}\times$  & \multicolumn{1}{c}{$0.95\times$} 
% & $1.26$ 
\\
& \multirow{2}{*}{30} & 
& 3330.22       & 4752.84       & \multicolumn{1}{c|}{5045.25} 
& 514.00        & 638.78        & \multicolumn{1}{c|}{672.07} 
& 0.03          & 0.03          & \multicolumn{1}{c}{$0.03$} \\
& & 
& $1.00\times$  & $1.43\times$  & \multicolumn{1}{c|}{$\mathbf{1.51}\times$} 
& $1.00\times$  & $1.24\times$  & \multicolumn{1}{c|}{$\mathbf{1.31}\times$}
& $1.00\times$  & $\mathbf{1.06}\times$  & \multicolumn{1}{c}{$1.03\times$} 
% & $1.46$ 
\\
& \multirow{2}{*}{50} & 
& 3269.34       & 6373.507      & \multicolumn{1}{c|}{7051.68} 
& 519.51        & 728.82        & \multicolumn{1}{c|}{783.18} 
& 0.03 & 0.04 & \multicolumn{1}{c}{$0.04$} \\
& & 
& $1.00\times$  & $1.95\times$  & \multicolumn{1}{c|}{$\mathbf{2.16}\times$} 
& $1.00\times$  & $1.40\times$  & \multicolumn{1}{c|}{$\mathbf{1.51}\times$} 
& $1.00\times$ & $1.28\times$ & \multicolumn{1}{c}{$\mathbf{1.30}\times$} 
% & $1.66$
\\ \\

\multirow{6}{*}{WSC}
& \multirow{2}{*}{15} & 
& 4287.24       & 13485.49    & \multicolumn{1}{c|}{38106.98} 
& 510.84        & 1008.59       & \multicolumn{1}{c|}{2454.89} 
& 0.04 & 0.07 & \multicolumn{1}{c}{$0.20$} \\
& & 
& $1.00\times$  & $3.15\times$  & \multicolumn{1}{c|}{$\mathbf{8.89}\times$} 
& $1.00\times$  & $1.97\times$  & \multicolumn{1}{c|}{$\mathbf{4.81}\times$} 
& $1.00\times$  & $2.02\times$ & \multicolumn{1}{c}{$\mathbf{5.51}\times$} 
% & $1.61$ 
\\
& \multirow{2}{*}{30} & 
& 4945.47       & 36984.44      & \multicolumn{1}{c|}{79786.57} 
& 573.78        & 2319.05       & \multicolumn{1}{c|}{5012.75} 
& 0.04          & 0.20          & \multicolumn{1}{c}{$0.46$} \\
& & 
& $1.00\times$  & $7.48\times$  & \multicolumn{1}{c|}{$\mathbf{16.13}\times$} 
& $1.00\times$  & $4.04\times$  & \multicolumn{1}{c|}{$\mathbf{8.74}\times$} 
& $1.00\times$  & $4.89\times$  & \multicolumn{1}{c}{$\mathbf{11.04}\times$} 
% & $1.46$ 
\\
& \multirow{2}{*}{50} & 
& 6002.68       & 81017.01      & \multicolumn{1}{c|}{159925.23} 
& 716.96        & 5093.42       & \multicolumn{1}{c|}{10192.41} 
& 0.05 & 0.46 & \multicolumn{1}{c}{$0.93$} \\
& & 
& $1.00\times$  & $13.50\times$ & \multicolumn{1}{c|}{$\mathbf{26.64}\times$} 
& $1.00\times$  & $7.10\times$  & \multicolumn{1}{c|}{$\mathbf{14.22}\times$} 
& $1.00\times$ & $10.16\times$ & \multicolumn{1}{c}{$\mathbf{20.56}\times$}  
% & $1.29$ 
\\
\multicolumn{12}{l}{} \\
\midrule

\multicolumn{12}{l}{\textit{\underline{WMT14/16 with \cite{ott2018scaling}}}}\vspace{3mm}\\
\multirow{2}{*}{En$\rightarrow$Fr} 
& \multirow{2}{*}{15} & 
& 9492.30       & 25772.89      & \multicolumn{1}{c|}{40975.78} 
& 1793.84       & 4961.56       & \multicolumn{1}{c|}{8494.36} 
& 0.10 & 0.24 & \multicolumn{1}{c}{$0.37$} \\
& & 
& $1.00\times$  & $2.72\times$  & \multicolumn{1}{c|}{$\mathbf{4.32}\times$} 
& $1.00\times$  & $2.77\times$  & \multicolumn{1}{c|}{$\mathbf{4.74}\times$} 
& $1.00\times$ & $2.51\times$ & \multicolumn{1}{c}{$\mathbf{3.89}\times$} 
% & $1.11$ 
\\

\multirow{2}{*}{En$\rightarrow$De} 
& \multirow{2}{*}{15} & 
& 8573.59       & 13293.51      & \multicolumn{1}{c|}{238677.16} 
& 1571.59       & 2476.18       & \multicolumn{1}{c|}{48446.29} 
& 0.09          & 0.13          & \multicolumn{1}{c}{$2.09$} \\
& & 
& $1.00\times$  & $1.55\times$ & \multicolumn{1}{c|}{$\mathbf{27.84}\times$} 
& $1.00\times$  & $1.58\times$ & \multicolumn{1}{c|}{$\mathbf{30.83}\times$} 
& $1.00\times$ & $1.46\times$ & \multicolumn{1}{c}{$\mathbf{24.18}\times$} 
% & $1.15$ 
\\
\multicolumn{12}{l}{} \\
\midrule

\multicolumn{12}{l}{\textit{\underline{WMT18 with \cite{edunov2018understanding}}}}\vspace{3mm}\\

\multirow{2}{*}{En$\rightarrow$De} 
& \multirow{2}{*}{15} & 
& 28393.97      & 38493.96      & \multicolumn{1}{c|}{874862.97} 
& 1624.05       & 2318.50       & \multicolumn{1}{c|}{49617.68} 
& 0.27          & 0.33          & \multicolumn{1}{c}{$7.25$} \\
& & 
& $1.00\times$  & $1.36\times$  & \multicolumn{1}{c|}{$\mathbf{30.81}\times$} 
& $1.00\times$  & $1.43\times$  & \multicolumn{1}{c|}{$\mathbf{30.55}\times$} 
& $1.00\times$ & $1.20\times$ & \multicolumn{1}{c}{$\mathbf{26.49}\times$} 
% & $1.16$ 
\\
\multicolumn{12}{l}{} \\
\midrule

\multicolumn{12}{l}{\textit{\underline{WMT19 with \cite{ng2019facebook}}}}\vspace{3mm}\\
\multirow{2}{*}{En$\rightarrow$Ru} 
& \multirow{2}{*}{15} & 
& 33181.43      & 91513.13      & \multicolumn{1}{c|}{876941.24} 
& 1897.19       & 5380.20       & \multicolumn{1}{c|}{47931.11} 
& 0.31          & 0.77          & \multicolumn{1}{c}{$7.19$} \\
& & 
& $1.00\times$  & $2.76\times$  & \multicolumn{1}{c|}{$\mathbf{26.43}\times$} 
& $1.00\times$  & $2.84\times$  & \multicolumn{1}{c|}{$\mathbf{25.26}\times$}
& $1.00\times$ & $2.46\times$ & \multicolumn{1}{c}{$\mathbf{22.85}\times$} 
% & $1.15$ 
\\
\multicolumn{12}{l}{} \\
\bottomrule
\end{tabularx}
\end{adjustbox}
\caption{Energy is reported in milli joules. 
We use the White-box GA attack to produce sponge examples and measure the performance on different platforms. The GPU readings are from NVML. GA was run for 1000 epochs with a pool size of 1000. A detailed explanation of the results is in \Cref{sec:eval:language}. Standard deviation for ASIC measurements are shown in \Cref{tab:nlp-ratios-stds}. }
\label{tab:nlp-bench-full}
\end{table*}

In this section, we look at the White-box GA attack (as explained in \Cref{sec:method:ga}) for generating sponge examples.
In this setup, we access to the parameters and run-time information of the neural networks. The GA optimisation relies on the estimated number of operations of the neural network inference.

\Cref{tab:nlp-bench-full} shows the energy consumption of different models in the presence of our generated White-box sponge examples.
For different input sequence sizes and a wide range of NLP tasks, we show the energy costs of sponge examples on both GPUs (GPU Energy) and the ASIC simulator (ASIC Energy).
We use natural, random and sponge to represent the energy measured on data from the evaluation dataset, randomly formed strings and sponge examples.
In addition, we also report the latency of running these samples on GPUs.
Due to the limitation of the ASIC simulator, we cannot have faithful time measurements and these numbers are not reported.

We have made several important observations:
\begin{itemize}
    \item The energy cost of sponge examples is always the highest on both GPUs and ASICs. In the best-case scenario for the attacker, sponge examples increase energy consumption by $26 \times$.
    \item Randomly generated samples are more energy-consuming than natural samples.
    \item When the task is quick to execute, sponge examples do not show big performance degradation in terms of GPU Time, but they increase latency significantly when the task takes more time (up to $30 \times$).
\end{itemize}

The main reason for performance degradation appears to be the increased dimension of the computation, as described in Algorithm \ref{alg:transformer}.
First, for a given input sequence size, the attack maximises the size of the post-tokenisation representation ($x_{\mathsf{tin}}$), exploiting the tokeniser and sub-word processing. Words with which the model is less familiar are represented inefficiently, forcing the network to do more work. Imagine holding an email conversation with an academic from another field, who uses specific technical terms. Every time an unfamiliar term appears you have to look it up in a dictionary or search engine. Second, the attack learns to maximise output sequence length, since this links directly to the computation cost.
Third, internal computation coupled with output sequence length and post-tokenisation length give a quadratic increase in energy consumption.
These reasons explain why sponge examples can significantly increase both energy and latency of language models. Do note that in this paper we use relatively small input sizes and in practice the effect will be a lot more pronounced for larger texts. Indeed as we later show in~\Cref{sec:azure}, in a Black-box setup with 50-character long text inputs an attack on Azure Language Translator caused $6000\times$ degradation.

Interestingly, we observe that randomly generated samples significantly reduce the performance of NLP tasks. This can be attributed to the fact that natural samples are efficiently encoded, whereas random and attack samples produce an unnecessarily long representation, meaning that random noise can be used as a scalable Black-box latency and energy attack tool. Otherwise put, many ML systems are vulnerable to simple barrage jamming.

It is also worth noting that the short execution of inference on GPUs makes it hard to provide an accurate measurement even with iterative runs. We further explain how this measurement is difficult due to a variety of hardware problems in~\Cref{sec:eval:cv}.

In the upcoming sections, we turn to Black-box variants of the attack based on energy and latency measurements of the individual samples. We mentioned previously that modern hardware optimises Operations per Watt (OPs/W), making sure that energy is only actively consumed when useful work is being done. In our experiments we see that the relationship between degradation factors of energy and time ranges between $1.15$ and $1.62$ (see~\Cref{tab:nlp-ratios} in Appendix), with energy scaling faster\footnote{Interestingly, we observe a net energy increase for the task if throttling happens. Although throttling decreases the running frequency and the voltage, it significantly increases the execution time so that the overall energy consumption has increased.}.

\subsection{Interactive Black-box Sponge Examples}
\label{sec:blind_attack}

In this section, we show the performance of the attacks running in an interactive Black-box manner against NLP tasks. 
In this setup, we launch the Black-box GA attack as described in \Cref{sec:method:ga}.
This interactive Black-box setup assumes that attackers cannot access to the neural network parameters but have the abilities to measure the energy or latency remotely. In addition, they can query the service as many times as they like, so there is no rate limiting.
We evaluate two Black-box attacks, and they use GPU Time and GPU Energy as the optimisation targets for the GA.
We also present results for a White-box GA attack in the third setup as a baseline, which is the same attack used in \Cref{sec:eval:language}.

\begin{figure*}[h]
    \centering
    \subfloat[][GPU Energy]{{\includegraphics[width=0.33\linewidth]{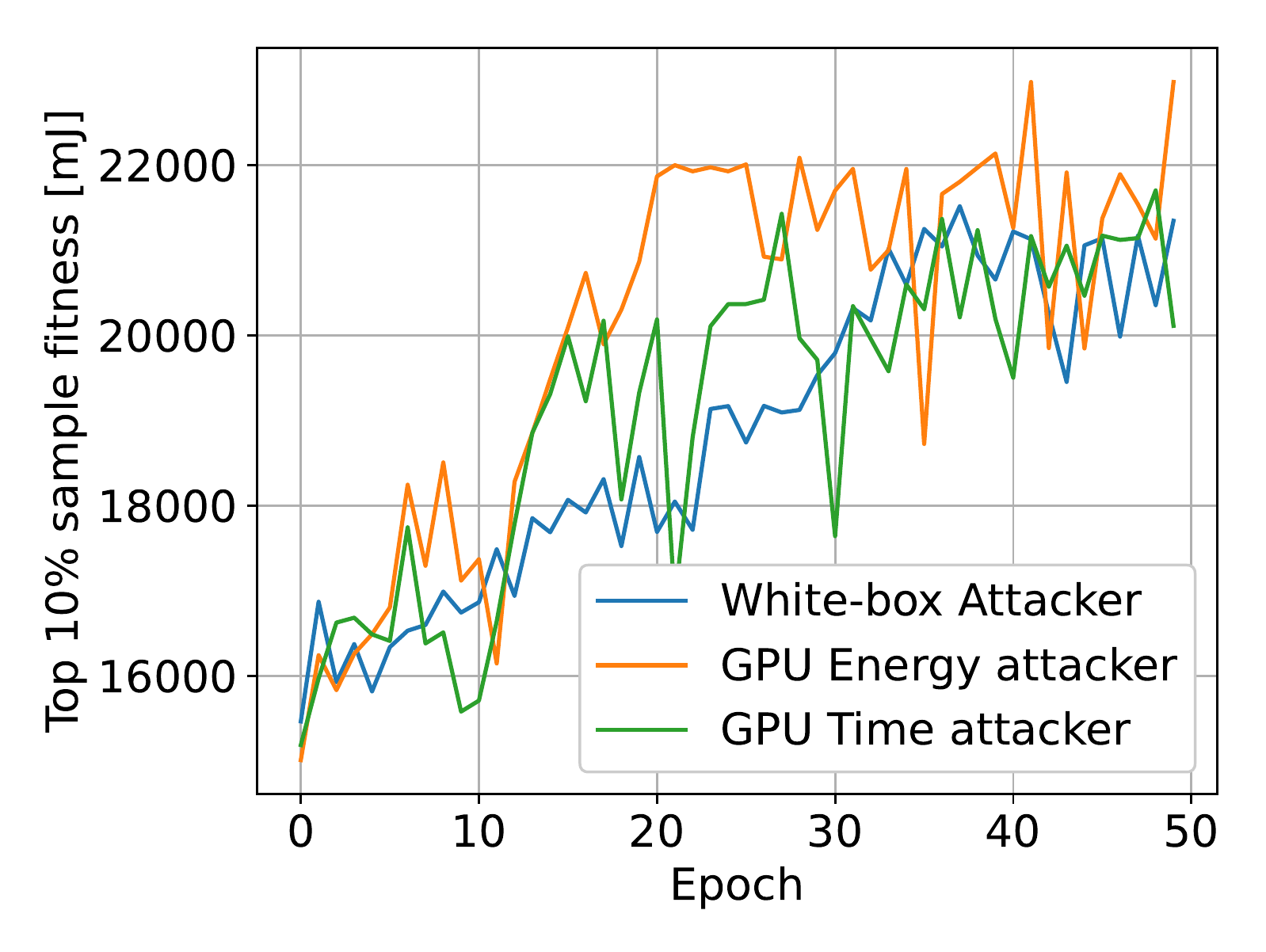}}}
    \subfloat[][GPU Time]{{\includegraphics[width=0.33\linewidth]{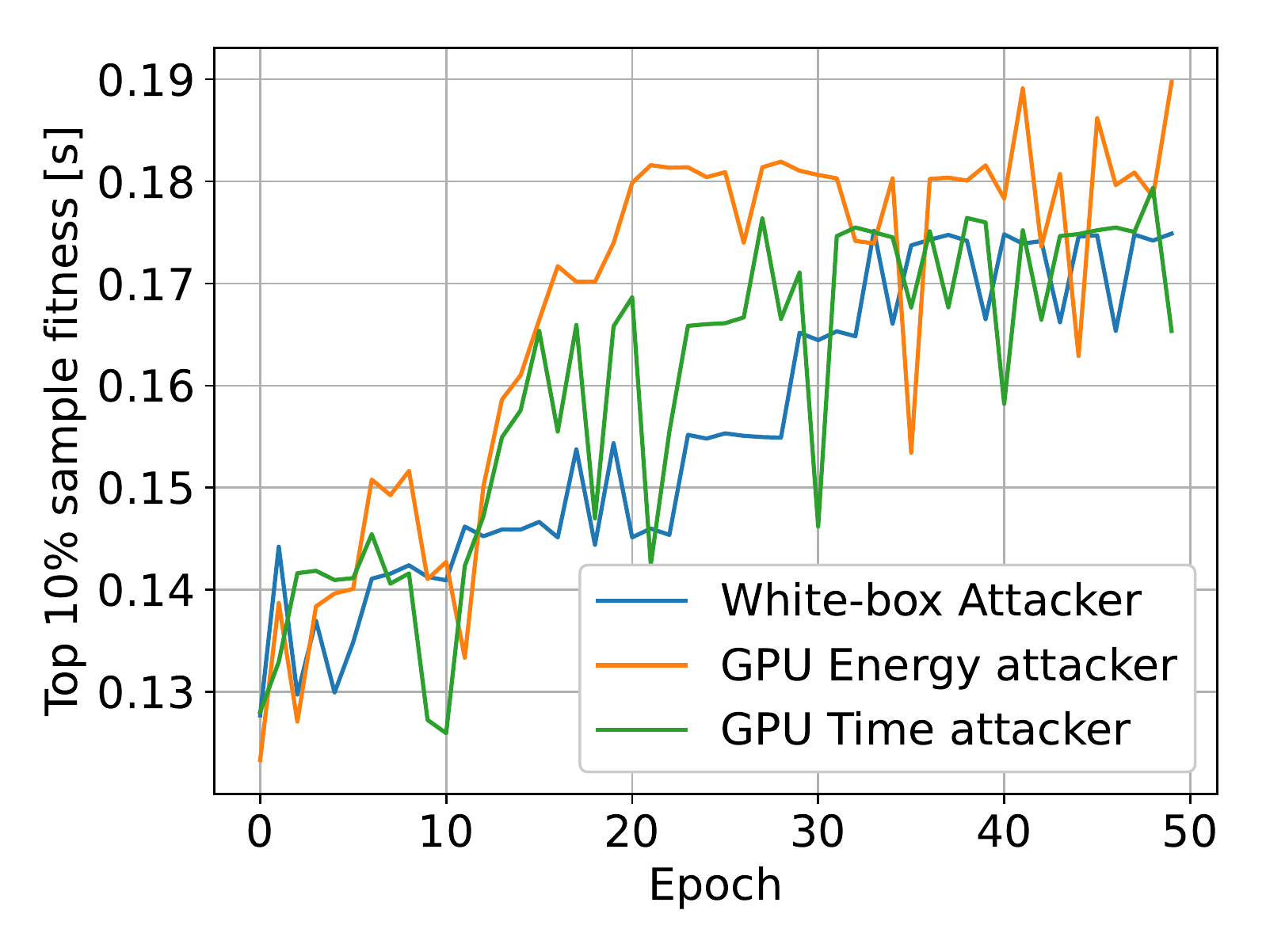}}}
    \subfloat[][ASIC Energy]{{\includegraphics[width=0.33\linewidth]{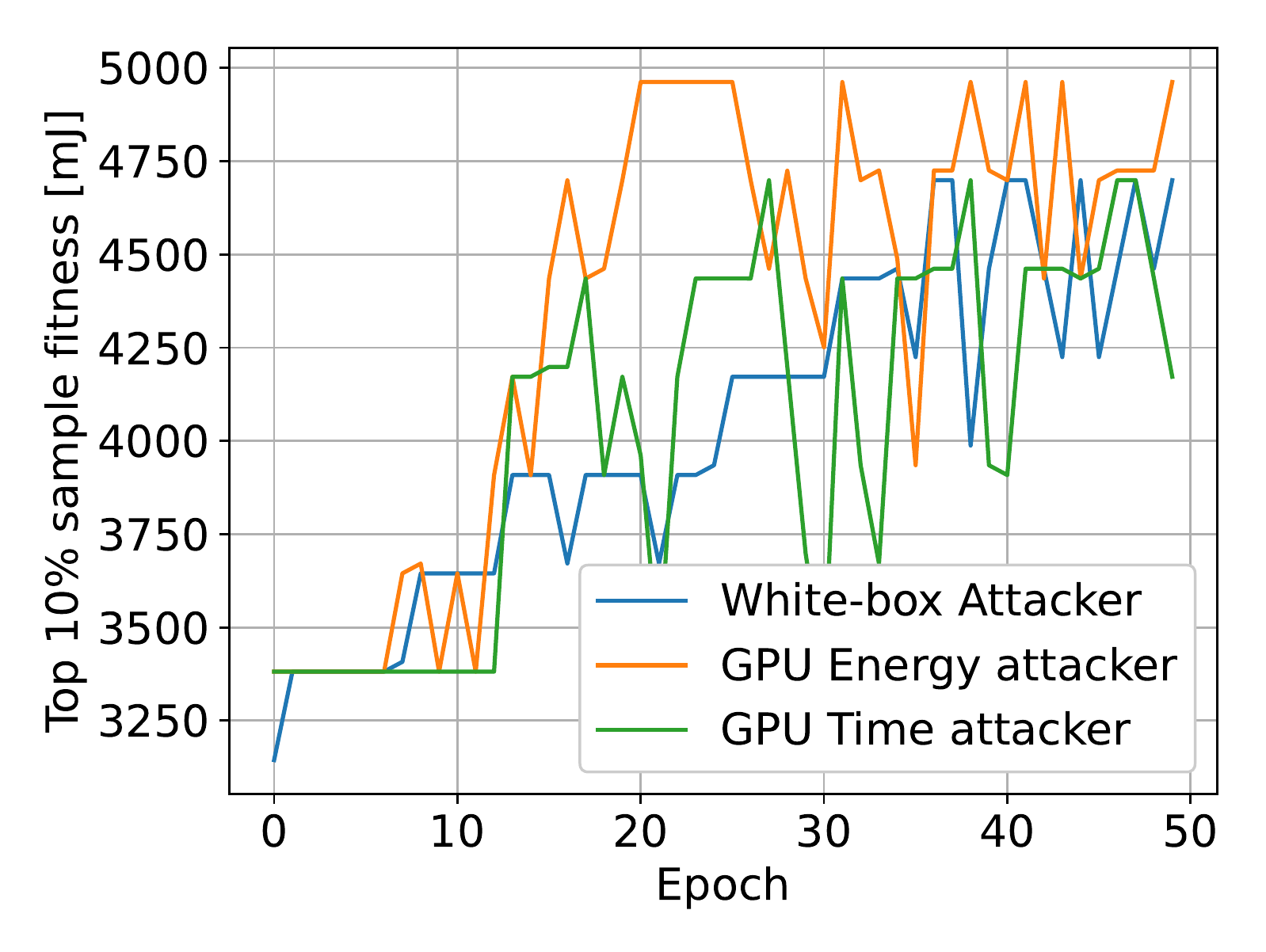}}}
    \caption{Black-box attack performance of sponge examples on different hardware metrics against English-to-French translation model~\cite{edunov2018understanding}. We show two Black-box attackers (GPU Energy and GPU Time attacker) and one White-box attacker, all using GA as the optimisation for finding sponge examples.}
    \label{fig:blackbox_ga}
\end{figure*}

\Cref{fig:blackbox_ga} shows sponge example performance against a WMT14 English-to-French Transformer-based translator with an input of size 15 and pool size of 1000. 
In \Cref{fig:blackbox_ga}, we use the name GPU Energy Attack, GPU Time Attacker and White-box Attacker to represent these different attacks.
In addition, we report measurements of every iteration of the GA for these different attackers on GPU Energy, GPU Time and ASIC Energy respectively. 
In \Cref{fig:blackbox_ga}, the legends represent attackers with different measurement proxies; and we show that these interactive Black-box attacks are transferable across hardware platforms and measurement proxies.
For instance, an attack targeting GPU Time transfers well when used to increase the energy cost of the ASIC simulator.

It can be seen that although the attackers have no knowledge of any neural network internals or datasets, they are able to successfully increase the energy and time costs. The experiment further highlights the difference between using time and energy as fitness functions. While time is noisy and depends on the current state of the hardware, energy remains relatively stable during the attack. As was explained previously, that can be attributed to the hardware switching its performance modes to keep the ratio of useful work to energy constant. 

\begin{table*}[!h]
\centering
\begin{adjustbox}{scale=1.15,center}
\begin{tabular}{@{}cllccccc@{}}
\toprule
                 &  & 
                &\textbf{ASIC}
                &\multicolumn{2}{c}{\textbf{GPU}} 
                &\multicolumn{2}{c}{\textbf{CPU}} 
                \\
                From & To &  &
                Energy [mJ]&
                Time [S] & 
                Energy [mJ]& 
                Time [S] &
                Energy [mJ] 
                \\
\midrule
\multicolumn{8}{l}{\textit{\underline{Black-box}}} \\
\multirow{10}{*}{$\text{WMT16}_{en\rightarrow de}$~\cite{ott2018scaling}}
& \multirow{3}{*}{
$\text{WMT14}_{en\rightarrow fr}$~\cite{ott2018scaling}} & Sponge 
& \multicolumn{1}{c|}{3648.219} 
& 0.174 & \multicolumn{1}{c|}{17251.000} 
& 1.048 & 51512.966 \\
& & Natural 
& \multicolumn{1}{c|}{1450.403} 
& 0.053 & \multicolumn{1}{c|}{6146.550} 
& 0.537 & 23610.145 \\
& & 
& \multicolumn{1}{c|}{$2.52\times$} 
& $\mathbf{3.27}\times$ & \multicolumn{1}{c|}{$\mathbf{2.81}\times$} 
& $1.95\times$ & $2.18\times$ \vspace{2mm}\\
& \multirow{3}{*}{
$\text{WMT18}_{en\rightarrow de}$~\cite{edunov2018understanding}} & Sponge 
& \multicolumn{1}{c|}{2909.245} 
& 0.414     & \multicolumn{1}{c|}{47723.500} 
& 3.199     & 181936.595 \\
& & Natural 
& \multicolumn{1}{c|}{1507.364}  
& 0.253     & \multicolumn{1}{c|}{27265.250} 
& 1.344     & 71714.201 \\
& & 
& \multicolumn{1}{c|}{$1.93\times$} 
& $1.64\times$  & \multicolumn{1}{c|}{$1.75\times$} 
& $\mathbf{2.38}\times$  & $\mathbf{2.54}\times$ \vspace{2mm}\\
& \multirow{3}{*}{
$\text{WMT19}_{en\rightarrow ru}$~\cite{ng2019WMT}} & Sponge 
& \multicolumn{1}{c|}{3875.365} 
& 0.652     & \multicolumn{1}{c|}{67183.100} 
& 4.409     & 247585.091 \\
& & Natural 
& \multicolumn{1}{c|}{1654.965} 
& 0.215     & \multicolumn{1}{c|}{25033.620} 
& 2.193     & 121210.376 \\
& & 
& \multicolumn{1}{c|}{$2.34\times$} 
& $\mathbf{3.03}\times$  & \multicolumn{1}{c|}{$\mathbf{2.68}\times$} 
& $2.01\times$  & $2.04\times$\\ \\
\midrule
\multicolumn{8}{l}{\textit{\underline{White-box}}} \\
\multirow{3}{*}{$\text{WMT16}_{en\rightarrow de}$~\cite{ott2018scaling}}& \multirow{3}{*}{
$\text{WMT16}_{en\rightarrow de}$~\cite{ott2018scaling}} & Sponge 
& \multicolumn{1}{c|}{48447.093} 
& 2.414         & \multicolumn{1}{c|}{260187.900} 
& 13.615        & 781758.680 \\
& & Natural 
& \multicolumn{1}{c|}{1360.118} 
& 0.056         & \multicolumn{1}{c|}{6355.620} 
& 0.520         & 23262.311 \\
& & 
& \multicolumn{1}{c|}{$35.62\times$} 
& $\mathbf{42.98}\times$ & \multicolumn{1}{c|}{$\mathbf{40.94}\times$} 
& $26.20\times$ & $33.61\times$ \\ \\
\bottomrule
\end{tabular}
\end{adjustbox}
\caption{Energy values are reported in milli Joules and time is reported in seconds. 
GA was run for 100 epochs with a pool size of 1000.
More results are available in Appendix. The first column shows source task that we generate sponge examples, and the second column shows the target task to launch these sponge examples. The performance of the sponge examples are evaluated on three hardware platforms (ASIC, GPU and CPU).}
\label{tab:bbx_hwtransfer}
\end{table*}

\subsection{Blind Black-box Sponge Examples and Hardware Transferability}
\label{sec:blackbox_hw_transfer}

In this section, we turn to the question of transferability across hardware and different models in a blind Black-box manner. 
As we've described in \Cref{sec:adversary}, in the blind Black-box setup the attacker blindly transfers the previously discovered sponge examples to the target. 
\Cref{tab:bbx_hwtransfer} shows the results across different models, tasks and hardware platforms. 
The first column is the source task that we used to produce sponge examples. We then later launch these sponge examples to the target tasks shown in the second column.
We report the performance of both sponge and natural examples on the targeting task in \Cref{tab:bbx_hwtransfer}.
Since the ASIC simulator is coarse-grained, it does not produce faithful execution time estimation, we thus only report the estimated energy cost.
In general, in \Cref{tab:bbx_hwtransfer}, we observe a significant increase in energy and time in comparison to natural samples on all hardware platforms. However, the blind Black-box attacks fail to achieve the same level of energy or latency degradation when taking the White-box case as a baseline. %We show the results for blind Black-box Sponge attacks on CV tasks in~\Cref{apdx:sec:cv_attack_transfer}.

%In \Cref{sec:blind_attack} we showed that an attacker can learn systems that it can interact with. Here, we assume that no interaction is possible and look at transferability instead. 

% 32FP: min = 2^-126
% 16FP: min = 2^-30
% 8FP: min = 2^-14
% Note for pedants: these are the smallest normalised numbers supported. Denormalised numbers also exist and can be smaller, but they're not very useful.

% \subsection{EEE Attack Hardware Transferbility}
% \label{sec:transferability_analysis}

% As attacks themselves do not particularly focus on specific hardware, but rather simulation of different hardware optimisation they exhibit a strong transferability effect. Furthermore, as a large part of the effect comes from inefficiencies in encoding those should carry a memory access cost everywhere.

\begin{figure*}[t]
    \centering
    \subfloat[][Requesting server measured]{{\includegraphics[width=0.47\linewidth]{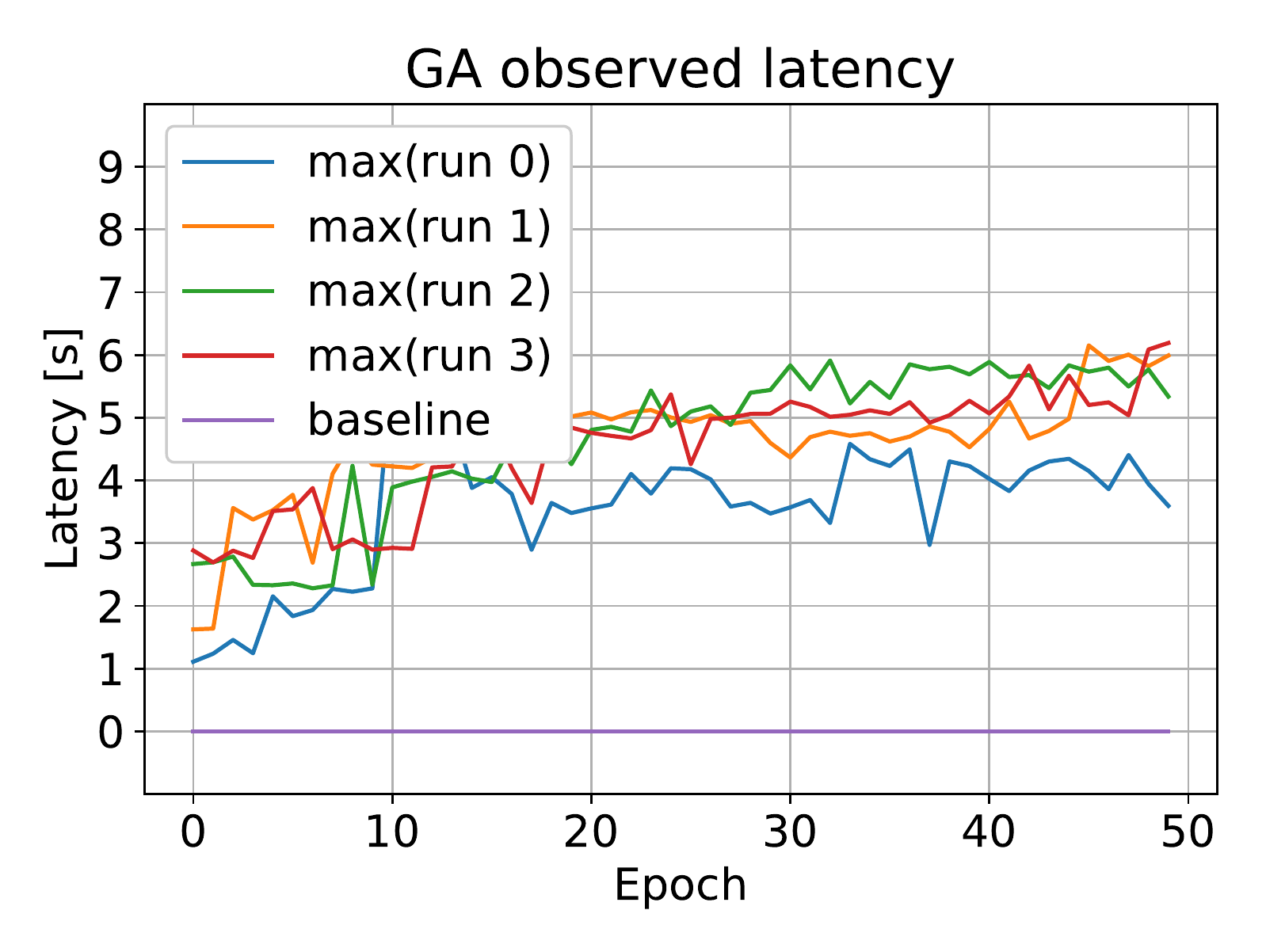}}}
    \qquad
    \subfloat[][Azure reported]{{\includegraphics[width=0.47\linewidth]{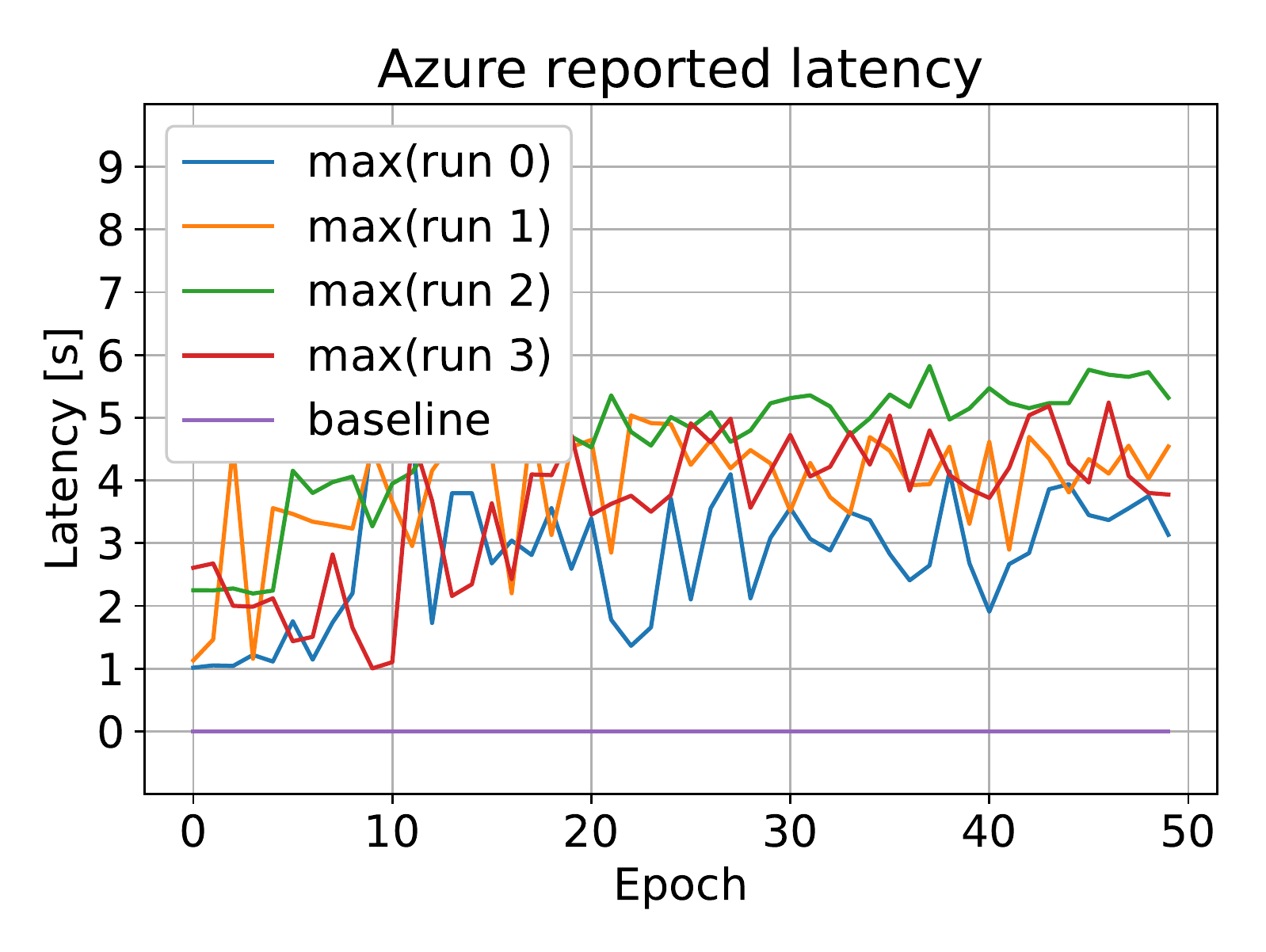}}}
    \caption{Maximum latency of the Microsoft Azure Translator model as is observed on the requesting server (a); reported by Azure servers (b). Azure servers were located on the same continent as the requesting server. Natural data mean baseline is at 1ms. We report multiple attack runs to show that the attack performs consistently with multiple restarts and the performance is not specific to the throttling of the user account. }
    \label{fig:azure_max_latency}
\end{figure*}

\subsection{A Case Study: Microsoft Azure Translation}
\label{sec:azure}
We evaluated sponge attack performance against an actually deployed service that is available on demand. We present a Black-box attack against this production system without any assumptions about its internals. As with the experiment setup in~\Cref{sec:blind_attack}, we interactively evolve the pool of samples in a Black-box setting using latency as a fitness function. Note that, in this case, observed sample fitness is noisy as it includes communication latency. That in turn makes it harder to perform the attack. 

We used the Microsoft Azure Translation system located on the same continent as the requesting server. We fixed the input to be 50 characters long, with a pool size of 500, and ran the attack for 50 epochs. We report four different attack runs, each running immediately when the previous attack finishes. The attacks are run sequentially, so Azure only translates a single sample at a time. It should be noted that we hold no assumptions about the actual system and do not possess any information about what architecture or dataset is used. Furthermore, we possess no information on whether Azure employs query-caching strategies or other optimisation techniques. 

\Cref{fig:azure_max_latency} shows the performance of the attack as observed by the requesting server and reported by Azure. The server's reported numbers are larger as they include additional noise from communication latency. It can be clearly seen that all four separate runs of GA were capable of converging to samples that were consuming considerably more time to process -- up to a maximum degradation factor of $6000\times$. Although we have no way of telling the amount of energy that was consumed by the computation, results from~\Cref{sec:eval_nlp} suggest that the energy consumption increase should also be in the range of thousands. Interestingly, we find that performance varies greatly within the pool during the attack. We suspect this is due to Azure's caching mechanism, where previously performing samples get almost constant time computation and the pool has to adapt quickly. For all individual runs, we see an up--down pattern that we do not observe with experiments on our own hardware. Interestingly, Azure translator assigns high confidence scores $>0.9$ to the sponge example predictions. 

Sponge examples against translators strongly resemble Denial-of-Service (DoS) attacks observed in the wild. DoS attacks aim to make computer systems unresponsive and unavailable via excess connections or data requests. Instead of overwhelming the victim's bandwidth as in the vast majority of DoS attacks~\cite{akamai_ddos_report}, we target the application layer. In particular, we target the most expensive parts of the translator to get an extraordinary amplification factor by sending specifically crafted requests. For example with Azure and an input of length 50, we were getting translated responses spanning thousands of characters. This finding bridges ML to the field of classic computer security and suggests that decades of experience with managing service-denial attacks can be applied here. 

\textbf{Ethics:} Having established the similarity of Sponge examples to DoS attacks, it is appropriate to discuss the ethics of the experiments. First, we paid for the translation service and used only legitimate and well-formed API calls. For experiments and testing, we performed around 200k queries. Second, to minimise the impact of sponge examples on Azure and CO$_2$ production, we chose relatively small input and pool sizes. Although the maximum input size that Azure accepts is 10000 characters per request, we used only 50. We expect that the impact of sponges can be further increased by running GA with a larger input size~\cite{azure_limits}.
Third, we ran the experiment at night in the data-center timezone, when it is easier to cool the servers and energy costs are lower. Fourth, to minimise the interaction of sponges between each other we executed a single sample at a time. Finally, we followed our standard responsible disclosure process: we notified Microsoft of the vulnerability of their Translator to sponge examples and shared a draft of the paper with the Microsoft Azure team. We want to stress that sponges are not specific to Microsoft Azure, and other ML-as-a-Service providers will be affected by the same attack. Microsoft Azure was chosen because of our experience with the Azure platform. Since the discovery, in a joint effort with Microsoft, sponge examples have been added to the MITRE attack framework for AI security\footnote{https://github.com/mitre/advmlthreatmatrix  
% \textbf{(warning: clicking on the link may reveal the identity of the authors of this paper)}
}.

\subsection{Section summary}
In this section, we demonstrate the effectiveness of sponge examples in different attack setups (White-box, Interactive Black-box and Blind Adversary). We consider a set of state-of-the-art sequence learning models, such as BERT \cite{devlin2018bert} and RoBERTa \cite{liu2019roberta}; and a wide variety of tasks. The performance of sponge examples are task-dependent and also model-dependent, however, all of the evaluated models and tasks show significant latency and energy increase when they are under attack. In addition, we demonstrate the transferability of sponge examples not only across hardware platforms but also across model architectures. Finally, we demonstrated how sponge examples can be generated in a Black-box manner on existing ML-as-a-service platforms, greatly increasing the response latency.

\begin{table*}[!h]
\centering
\begin{adjustbox}{scale=1.1,center}
\begin{tabular}{@{}lcccccccc@{}}
\toprule
& 
& \multicolumn{2}{c}{\textbf{Energy}}
& \multicolumn{3}{c}{\textbf{Density}}
\\
& 
& ASIC Energy [mJ] 
& Energy ratio 
& Post-ReLU  
& Overall
& Maximum \\
\midrule
\multicolumn{6}{l}{\textit{\underline{ImageNet}}} \vspace{3mm}\\

\multirow{4}{*}{ResNet-18} 
& Sponge LBFGS 
& 53.359 $\pm$ 0.004 &\multicolumn{1}{c|}{0.899}      & 0.685     & \textbf{0.896}     & \multirow{4}{*}{0.981} \\
& Sponge 
& 51.816 $\pm$ 0.271 & \multicolumn{1}{c|}{0.873}     & 0.599     & 0.869     &   \\
& Natural 
& 51.745 $\pm$ 0.506 & \multicolumn{1}{c|}{0.871}     & 0.596     & 0.869     &   \\
& Random 
& 49.685 $\pm$ 0.008  & \multicolumn{1}{c|}{0.837}     & 0.480     & 0.834     &  \vspace{2mm} \\
\multirow{4}{*}{ResNet-50}
& Sponge LBFGS 
& 164.727 $\pm$ 0.062 & \multicolumn{1}{c|}{0.863}       & 0.619     & \textbf{0.885} & \multirow{4}{*}{0.998} \\
& Sponge 
& 160.887 $\pm$ 0.609 & \multicolumn{1}{c|}{0.843}      & 0.562 & 0.868 & \\
& Natural 
& 160.573 $\pm$ 1.399 & \multicolumn{1}{c|}{0.842}      & 0.572 & 0.867 &  \\
& Random 
& 155.819 $\pm$ 0.016 & \multicolumn{1}{c|}{0.817}      & 0.483 & 0.845 & \vspace{2mm} \\

\multirow{4}{*}{ResNet-101} 
& Sponge LBFGS 
& 258.526 $\pm$ 0.028 & \multicolumn{1}{c|}{0.857}      & 0.597 & \textbf{0.873} & \multirow{4}{*}{0.994}  \\
& Sponge 
& 254.182 $\pm$ 0.561 & \multicolumn{1}{c|}{0.842}      & 0.556 & 0.861 &   \\
& Natural 
& 253.004 $\pm$ 1.345 & \multicolumn{1}{c|}{0.839}      & 0.545 & 0.857 &   \\
& Random 
& 249.026 $\pm$ 0.036 & \multicolumn{1}{c|}{0.825}      & 0.507 & 0.846 & \vspace{5mm} \\

\multirow{4}{*}{DenseNet-121}
& Sponge LBFGS 
& 152.595 $\pm$ 0.050 & \multicolumn{1}{c|}{0.783} & 0.571 & \textbf{0.826} & \multirow{4}{*}{0.829} \\
& Sponge 
& 149.564 $\pm$ 0.502 & \multicolumn{1}{c|}{0.767} & 0.540 & 0.814 & \\
& Natural 
& 147.247 $\pm$ 1.199 & \multicolumn{1}{c|}{0.755} & 0.523 & 0.804 &  \\
& Random 
& 144.366 $\pm$ 0.036 & \multicolumn{1}{c|}{0.741} & 0.487 & 0.792 & \vspace{2mm} \\

\multirow{4}{*}{DenseNet-161} 
& Sponge LBFGS 
& 288.427 $\pm$ 0.087 & \multicolumn{1}{c|}{0.726} & 0.435 & \textbf{0.764} & \multirow{4}{*}{0.811}  \\
& Sponge 
& 287.153 $\pm$ 0.575 & \multicolumn{1}{c|}{0.723} & 0.429 & 0.761 &   \\
& Natural 
& 282.296 $\pm$ 2.237 & \multicolumn{1}{c|}{0.711} & 0.404 & 0.751 &   \\
& Random 
& 279.270 $\pm$ 0.065 & \multicolumn{1}{c|}{0.703} & 0.387 & 0.744 &  \vspace{2mm} \\

\multirow{4}{*}{DenseNet-201} 
& Sponge LBFGS 
& 237.745 $\pm$ 0.156 & \multicolumn{1}{c|}{0.756} & 0.505 & 0.788 & \multirow{4}{*}{0.863} \\
& Sponge 
& 239.845 $\pm$ 0.522 & \multicolumn{1}{c|}{0.763} & 0.519 & \textbf{0.794} &  \\
& Natural 
& 234.886 $\pm$ 1.708 & \multicolumn{1}{c|}{0.747} & 0.487 & 0.781 &  \\
& Random 
& 233.699 $\pm$ 0.098 & \multicolumn{1}{c|}{0.743} & 0.479 & 0.777 & \vspace{5mm} \\

\multirow{4}{*}{MobileNet v2}
& Sponge LBFGS
& 87.511 $\pm$ 0.011 & \multicolumn{1}{c|}{0.844}  & 0.692 & \textbf{0.890} & \multirow{4}{*}{0.996} \\
& Sponge 
& 84.513 $\pm$ 0.386 & \multicolumn{1}{c|}{0.815}  & 0.645 & 0.868 & \\
& Natural 
& 85.077 $\pm$ 0.683 & \multicolumn{1}{c|}{0.821}  & 0.646 & 0.873 & \\
& Random 
& 80.807 $\pm$ 0.022 & \multicolumn{1}{c|}{0.779}  & 0.567 & 0.844 & \\ \\
\bottomrule
\end{tabular}
\end{adjustbox}
\caption{
We report the performance of two White-box attacks, Sponge and Sponge LBFGS against a number of computer vision benchmarks. They are optimised using the the GA and LBFGS respectively for finding sponge examples. We show the energy readings from the ASIC simulator and the Energy ratio. The Energy ratio is a ratio between the estimated energy of an ASIC optimised for sparse matrix multiplication and an ASIC without such optimisations. To further illustrate the internals of neural networks, we show data densities that are post-ReLU, across the entire neural network, and also the maximum possible density calculated using interval bound propagation (IBP). Details are described in \Cref{sec:eval:cv}.}
\label{tab:cv-bench-full}
\end{table*}

\begin{figure*}[h]
    \centering
    \subfloat[][Sponge density - Normal density]{{\includegraphics[width=0.5\linewidth]{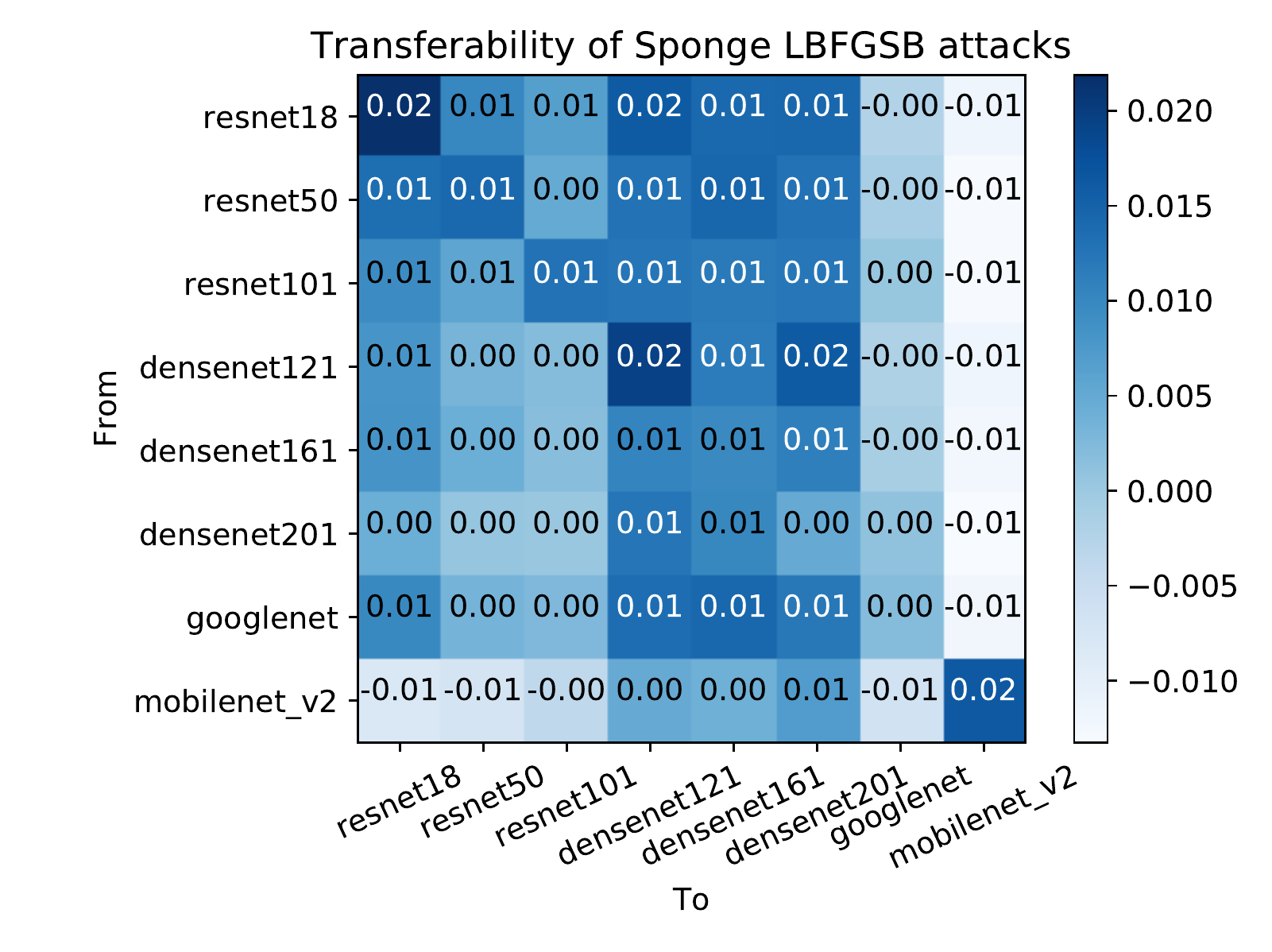}}}
    \subfloat[][Sponge density - Random density]{{\includegraphics[width=0.5\linewidth]{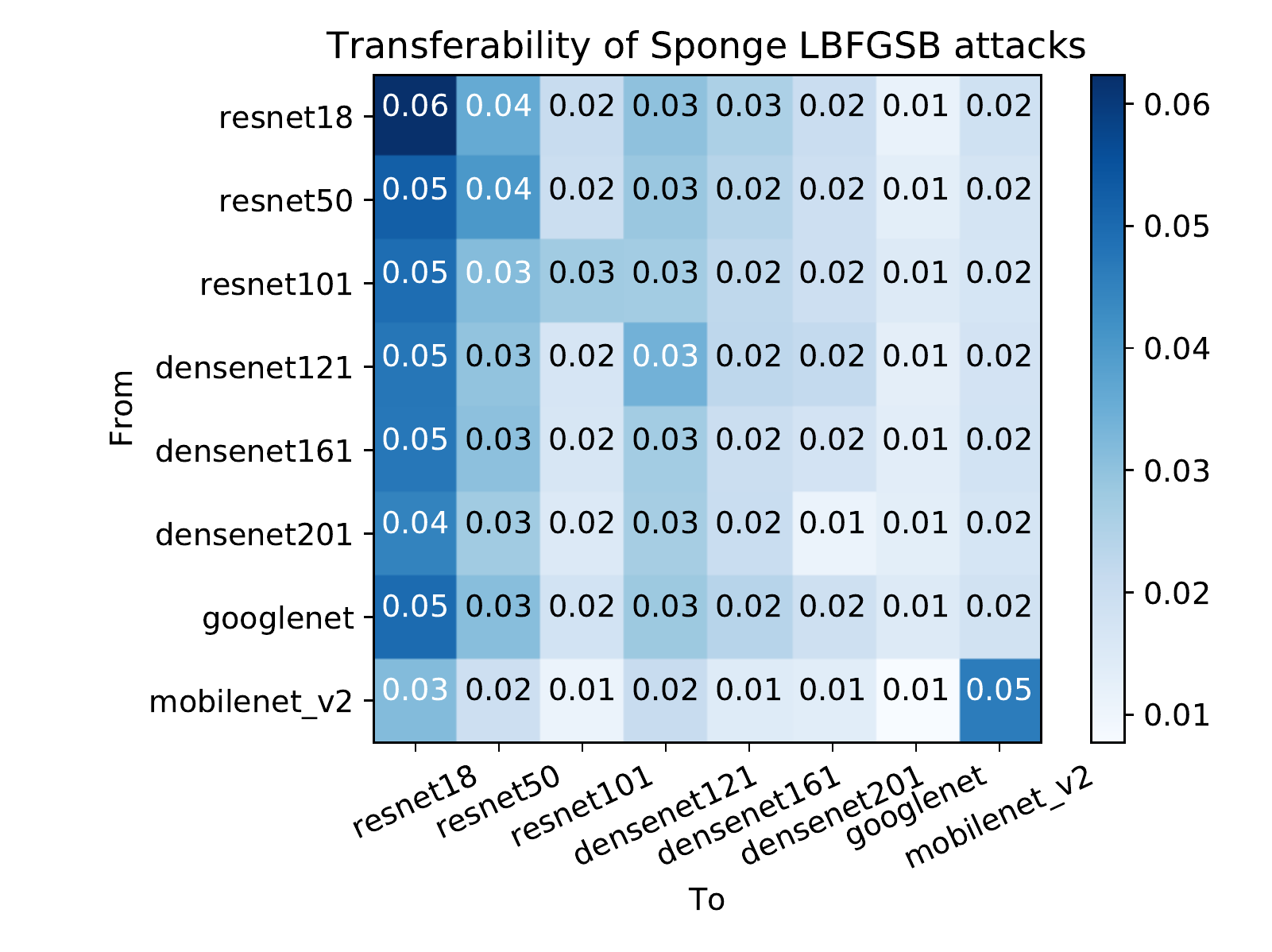}}}
    \caption{Transferability of sponge examples across different computer vision benchmarks.}
    \label{fig:transferability}
\end{figure*}

\begin{figure*}[h]
    \centering
    \subfloat[ResNet family]{{\includegraphics[width=0.33\linewidth]{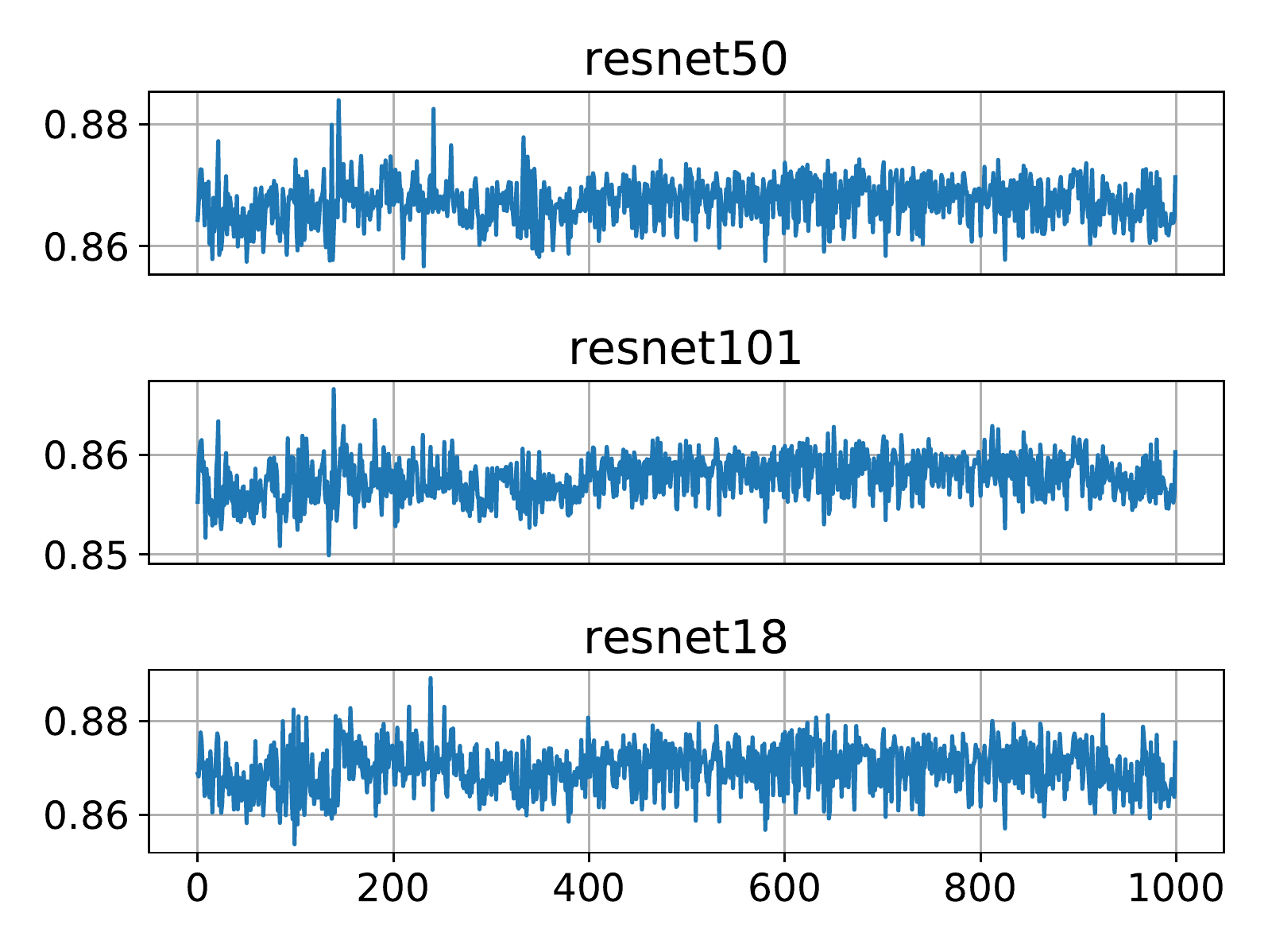}}}
    \subfloat[DenseNet family]{{\includegraphics[width=0.33\linewidth]{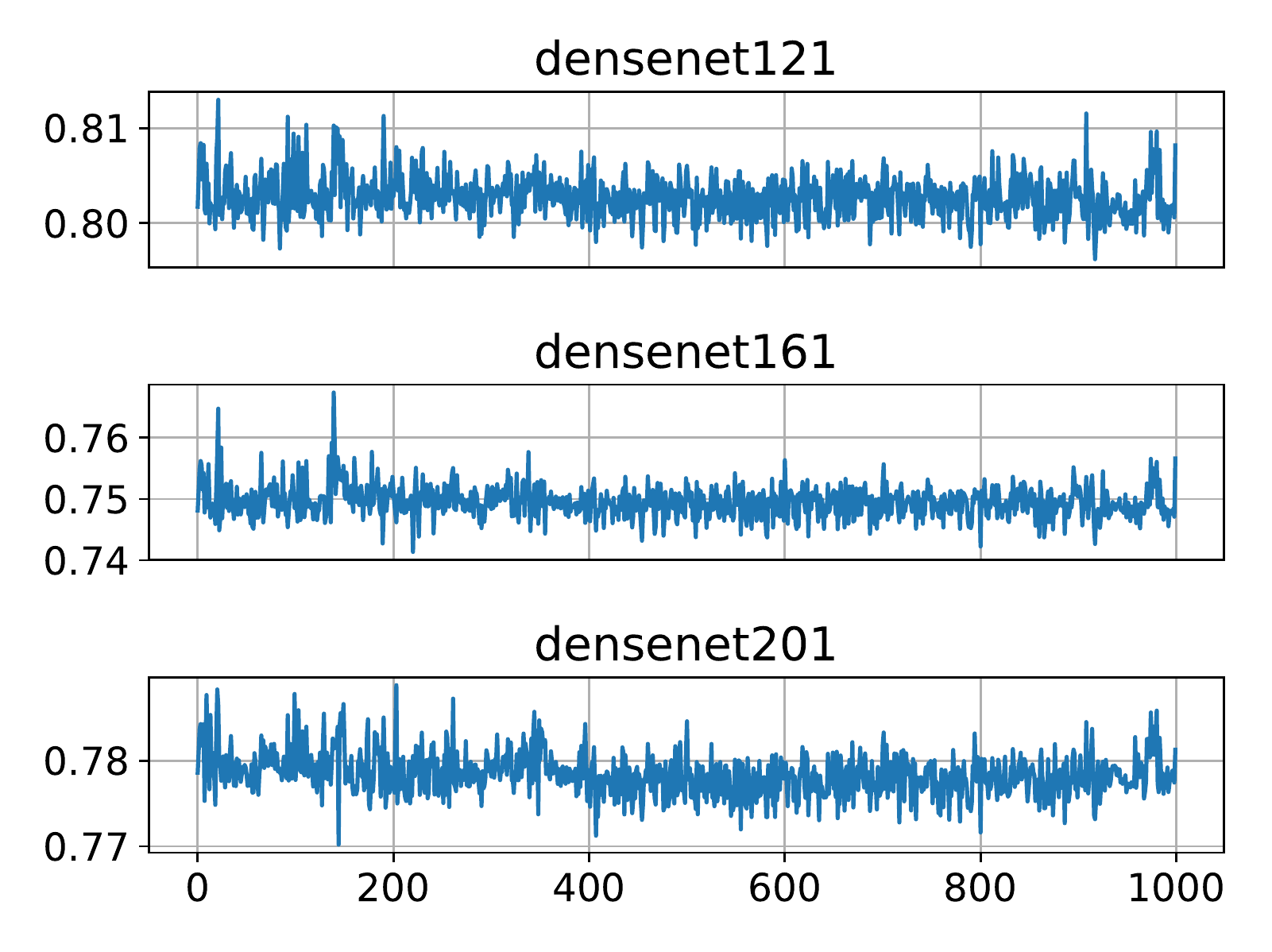}}}
    \subfloat[Other networks]{{\includegraphics[width=0.33\linewidth]{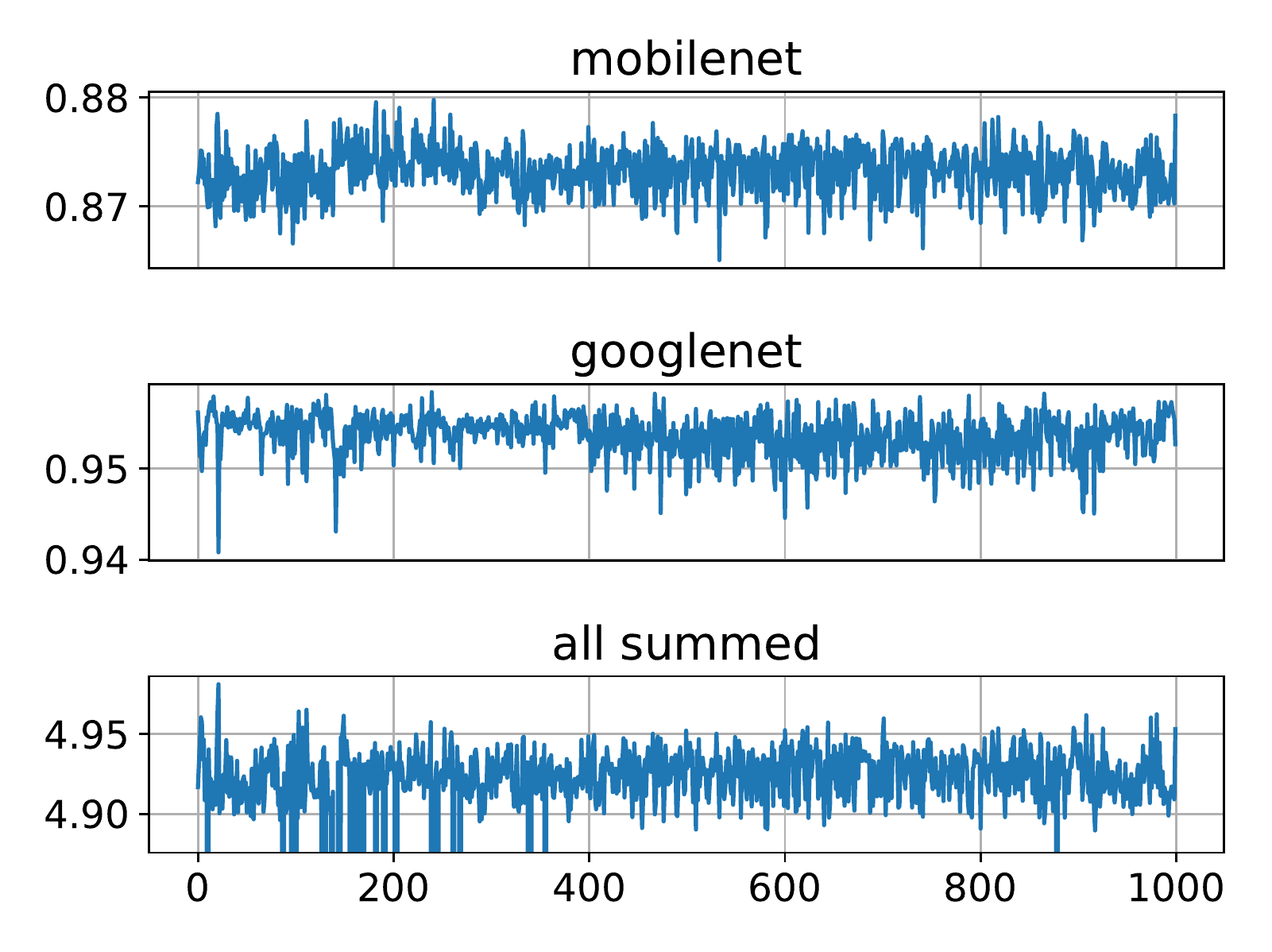}}}
    \caption{Class-wise average densities of natural samples from the ImageNet validation dataset. Some classes are a lot more densely represented internally than others. X-axis shows the class numbers, whereas Y-axis shows densities. }
    
    \label{fig:class_transferability}
\end{figure*}

\section{Sponge Examples on Vision Models}
\subsection{Models and Datasets}
We evaluate the sponge example attack on a range of vision models provided in the TorchVision library. We show the performance of ResNet-18, ResNet-50 and ResNet-101 \cite{he2016deep}, DenseNet-121, DenseNet-161, DenseNet-201 \cite{huang2017densely},
% GoogleNet \cite{szegedy2015going}
and MobileNet-V2 \cite{sandler2018mobilenetv2}. 
Networks span a range of sizes from 3.4M parameters (MobileNet-V2) to 49M (ResNet-101). The considered networks also have a relatively large architectural diversity, where MobileNet-V2 is designed to run on modern battery-powered mobile devices.
All of the networks classify a canonical computer vision task -- ImageNet-2017, since the ImageNet challenge serves as a golden baseline in the computer vision community.

\subsection{White-Box Sponge Examples}
\label{sec:eval:cv}

Following objectives in \Cref{sec:hypo1} and \Cref{sec:hypo2}, we can increase energy consumption by increasing either computation dimension or data density.
Although theoretically we can provide larger images to increase the computation dimension for computer vision networks, very few modern networks currently deal with dynamic input or output.
Usually preprocessing normalizes variable-sized images to a pre-defined size by either cropping or scaling.
% For datasets with variable-sized inputs it is common to assume a deterministic up-sampling or down-sampling component.
Therefore, for computer vision models, we focus on increasing energy and latency via data density.

\Cref{tab:cv-bench-full} shows the performance of sponge examples on CV models, and we focused on using White-box attacks to maximise energy consumption.
We use both the White-box GA and White-box L-BFGS to generate sponge examples (named sponge and sponge LBFGS in \Cref{tab:cv-bench-full}).
Since the energy consumption is lower per inference, it is challenging to get a true measurement of energy given the interference of the GPU's hardware temperature control, and that energy inspection tools lack the resolution.
We then show the ASIC Energy readings and the Energy Ratio in the first two columns.
The Energy Ratio term refers to the cost on an ASIC with data sparsity optimisations compared to the cost on an ASIC without any optimisations.
We considered data sparsity optimisations including compressed DRAM accesses, zero-skipping multiplications. These optimisation techniques are widely adopted in many proposed ASIC accelerators \cite{han2016eie,kim2017novel, parashar2017scnn}, and there are now real implementations of these techniques in hardware.
We then further look at the internals of neural networks and show how their data density is changing with different types of samples. 
We calculate the theoretical upper bounds of data density using Interval Bound Propagation (IBP)~\cite{gowal2018effectiveness}.
Although originally developed for certifiable robustness, we adopt the technique to look at internal network bounds that only take value $0$ (\ie~$\text{lower bound} = \text{upper bound} = 0$) for the whole natural image range\footnote{Note that we assume full floating point precision here. In practice, emerging hardware often uses much lower quantization which will result in a lower maximum data density.}.
We also look at data densities after the ReLU function (Post-ReLU) and the overall densities.
The results for density and energy suggest that both attacks can successfully generate sponge examples that are marginally more expensive in terms of energy. 
To be precise, we were able to get a $1-3\%$ increase in energy consumption when compared to natural samples. Interestingly we observe that more of the density impact comes in the first few layers. 
To better understand the difference in performance please refer to~\Cref{apdx:sec:understanding}.
We show a statistical analysis across a wide range of CV models and describe the difficulties of precisely showing performance on CPUs and GPUs in \Cref{apdx:sec:understanding}.

For computer vision models,
we also find that different architectures will have similar class-wise computation densities and sponge examples can increase densities across model architectures. 
% We present the details of the transferability of CV models in \Cref{apdx:sec:cv_attack_transfer} in Appendix.

\subsection{Transferability of Attacks}
\label{apdx:sec:cv_attack_transfer}

We observe that sponge examples are transferable and can be used to launch a blind Black-box attack. 
\Cref{fig:transferability} shows the density difference of transferred sponge samples. For all networks but one (MobileNet), the sponge samples increased the internal data density despite not having any knowledge of what natural samples look like or any architectural and parameter information of the targeted neural networks. All of the sponge samples outperformed random noise, suggesting that sponge samples target specific features of the data set and can be applied in a blind Black-box fashion.

\subsection{Class-Wise Natural Data Density}
\label{apdx:sec:class_transferability}

\Cref{fig:class_transferability} shows the densities of natural samples from the ImageNet dataset. 
On the horizontal axis, we show the 1000 classes of ImageNet and the vertical axis displays the run-time data densities for samples in that class.
It can be clearly seen that there are per-class similarities between data densities of natural samples. These are particularly pronounced within ResNet and DenseNet architectures hinting that similar architectures will learn similar features so that samples of the same class have similar run-time densities across architectures. Finally, \Cref{fig:class_transferability}.c shows the summed per-class densities across all of the tested networks. There are classes that are consistently more dense than others.
This test is helping us to summarise that, in computer vision tasks, there exist natural samples that are producing more computation because of the increased data densities. This intrinsic property suggests that an adversary may send \textit{natural} samples resulting in higher activation density to drain the energy of targeted devices.

\subsection{Section summary}
In this section, we report the results of sponge examples on computer vision models. We observe that sponge examples can successfully decrease the run-time sparsity of CV models, thus generating marginally more energy-consuming samples for ASICs that utilise data sparsity. The generated hardware differences are too small to be reliably observed on GPUs, however, we show that the GPU energy readings are statistically different between normal and sponge samples in~\Cref{apdx:sec:measuring_cv}. The computation of CNN inference is more structured and normally only handles fixed-sized inputs. This structured computation flow provides fewer opportunities for sponge examples, and only hardware devices utilising fine-grained sparsity are vulnerable to sponge attacks.

\section{Discussion}

\subsection{Lessons from Sponge Examples}

In this paper, we demonstrated novel service-denial attacks on ML components. In their current form, our attacks work best against NLP models, whose internal complexity makes domain-specific optimisations necessary. We showed that our attacks can also target hardware optimisations, suggesting that the capable attacker will always be capable of exploiting different optimisations across the stack. 

Our attacks will have a significant impact on how future ML pipelines will be deployed. The integration of different ML components can only lead to higher complexity, which will in turn be even more vulnerable to the sponge attacks described here. They may lead to ML deadlocks or livelocks, of which another precursor may be semi-trained RL agents that forever walk in circles. 

The attacks presented in this paper assume that a single sample is processed at a time. This enables simple demonstrations but these are only a starting point. More complex attacks could involve samples that interact with each other. Indeed, in the world of Federated Learning, aggregators that experience delays in presence of network failure are likely to find this leads to significant increases in overall latency. Coordinated attacks on federated systems are a natural frontier for future research.

Our attacks also used two main optimisation strategies, but others can be tried. In our attack on the Microsoft Azure Translator, it appears that a caching mechanism was making previously potent samples perform poorly in subsequent runs, but we still managed, using a genetic algorithm, to find powerful sponge examples. 

Our attacks were used offensively in this paper, but such examples should also be used routinely in safety testing. Worst-case examples may also turn up by happenstance, so it is prudent to use adversarial techniques to find them in advance. Furthermore, our methodology can be used to automatically discover timing side-channels and other latent dependencies between interacting components in both ML and traditional processing pipelines.

Finally, sponge examples show that commonly deployed API rate limiting defences are not enough to protect the availability of the underlying machine learning system. Indeed, the attacker can use sponges to increase consumption of the overall system per sample without increasing the rate at which the system is queried. 

\subsection{Defending against Sponge Examples}

Sponge examples can be found by adversaries with limited knowledge and capabilities, making the threat realistic. We further showed that sponge examples work against a deployed system that is available on demand. We now propose a simple defence to preserve the availability of hardware accelerators in the presence of sponge examples. 

In~\Cref{tab:nlp-bench-full}, we observe a large energy gap between natural examples and random or sponge examples. 
We propose that before deploying the model, natural examples get profiled to measure the time or energy cost of inference. The defender can then fix a~\textit{cut-off} threshold. This way, the maximum consumption of energy per inference run is controlled and sponge examples will simply result in an error message. In this way, their impact on availability can be bounded. 

This will often be sufficient to deal with the case where the threat model is battery drainage. Where the threat is a jamming attack on real-time performance, as with the vision system of an autonomous vehicle, the system will need to be designed for worst-case performance, and if need be a fallback driving mechanism should be provided. Current draft safety standards call for a self-driving car that becomes confused to slow to a stop in the same lane while alerting the human driver. This may be a good example of the appropriate response to an attack.

No single solution can tackle all of the possible abuse cases where an attacker can introduce errors into a machine-learning system. Depending on the setup both defence~\cite{goodfellow2014explaining,wong2020fast} and detection~\cite{chen2020stateful,shumailov2020ctt} mechanisms may be required. That problem space may be as large as the human conflict itself. At the level of technical research, serious work is needed to assess what impact different hardware platforms (\eg~TPUs that do not exploit sparsity) have on susceptibility to sponge examples. Above all, it is vital to take a whole system view when engineering for security or safety; to consider what threats and hazards are realistic; and to provide appropriate defences or mitigation. In the case of attacks that cannot be prevented, the optimal strategy may often be attack detection. 
% In our experiments we saw that hardware plays a large role and has a significant impact on performance degradation.

% \subsection{Reproducibility}

% It should be noted that the performance of our attacks will vary greatly across different hardware platforms. When running experiments in a black-box setup on two servers with similar configurations we found the energy and latency varied by up to a factor of 10. To help reproducibility, we will release the Sponge examples we found, the attack code-base we used and the ASIC simulator. 

\subsection{Energy and Machine Learning}
\label{sec:related:carbon}

Most of the prior research on the carbon footprint of machine learning focuses on the energy required to train large neural network models and its contribution to carbon emissions~\cite{henderson2020towards,lacoste2019quantifying,strubell2019energy}. This work shows that we need to study energy use at small scales as well as large. As with side-channel attacks on cryptographic systems, the fine-grained energy consumption of neural networks is a function of the inputs. In this case, the main consequence is not leakage of confidential information but a denial-of-service attack. 

First, sponge examples can aim to drain a device's batteries; the operations and memory access in inference account for around a third of the work done during a complete backpropagation step, but inference happens at a much higher frequency and scale compared to training once a model is deployed. Our research characterizes the worst-case energy consumption of inference. This is particularly pronounced with natural-language processing tasks, where the worst case can take dozens of times more energy than the average case.

Second, the sponge examples found by our attacks can be used in a targeted way to cause an embedded system to fall short of its performance goal. In the case of a machine-vision system in an autonomous vehicle, this might enable an attacker to confuse the scene understanding mechanisms and crash the vehicle; in the case of a missile guided by a neural network target tracker, a sponge example might break the tracking lock. The lesson is that system engineers must think about adversarial worst-case performance and test it carefully.

% \subsection{Reproducibility}
% We find that it is hard to measure energy in a reproducible fashion, the numbers are always changing. Furthermore, the actual platform state makes a lot of difference. 

\section{Reproducibility}

It should be noted that the performance of our attacks will vary greatly across different hardware platforms and even weather outside. When running experiments in a Black-box setup on two servers with similar configurations in some cases we found the energy and latency varied by up to a factor of 10. To help reproducibility, we release the sponge examples we found, the attack code-base we used and the ASIC simulator\footnote{\url{https://github.com/iliaishacked/sponge_examples}}. 

\section{Conclusion}

We introduced energy-latency attacks, which enable an adversary to increase the latency and energy consumption of ML systems to deny service. Our attacks use specially-crafted sponge examples and are effective against deep neural networks in a spectrum of threat models that realistically capture current deployments of ML -- whether as a service or on edge devices. They can be mounted by adversaries whose access varies from total to none at all. As proof of concept, we showed that we can slow down translation in Microsoft Azure by a factor of several thousand. Our work demonstrates the need for careful worst-case analysis of the latency and energy consumption of computational systems that use deep learning mechanisms.

\section*{Acknowledgment}

We thank the reviewers for their insightful feedback. We want to explicitly thank Nicholas Carlini, Florian Tramèr, Adelin Travers, Varun Chandrasekaran and Nicholas Boucher for their help and comments. This work was supported by CIFAR (through a Canada CIFAR AI Chair), by EPSRC, by Apple, by Bosch Forschungsstiftung im Stifterverband, by NSERC, and by a gift from Microsoft. We also thank the Vector Institute's sponsors.

\bibliographystyle{unsrt}
\bibliography{bibliography}

\begin{thebibliography}{10}

\bibitem{biggio2013evasion}
Battista Biggio, Igino Corona, Davide Maiorca, Blaine Nelson, Nedim
  {\v{S}}rndi{\'c}, Pavel Laskov, Giorgio Giacinto, and Fabio Roli.
\newblock Evasion attacks against machine learning at test time.
\newblock In {\em Joint European conference on machine learning and knowledge
  discovery in databases}, pages 387--402. Springer, 2013.

\bibitem{szegedy2013intriguing}
Christian Szegedy, Wojciech Zaremba, Ilya Sutskever, Joan Bruna, Dumitru Erhan,
  Ian Goodfellow, and Rob Fergus.
\newblock Intriguing properties of neural networks.
\newblock {\em arXiv preprint arXiv:1312.6199}, 2013.

\bibitem{nelson2008exploiting}
Blaine Nelson, Marco Barreno, Fuching~Jack Chi, Anthony~D Joseph, Benjamin~IP
  Rubinstein, Udam Saini, Charles~A Sutton, J~Doug Tygar, and Kai Xia.
\newblock Exploiting machine learning to subvert your spam filter.
\newblock {\em LEET}, 8:1--9, 2008.

\bibitem{jagielski2018manipulating}
Matthew Jagielski, Alina Oprea, Battista Biggio, Chang Liu, Cristina
  Nita-Rotaru, and Bo~Li.
\newblock Manipulating machine learning: Poisoning attacks and countermeasures
  for regression learning.
\newblock In {\em 2018 IEEE Symposium on Security and Privacy (SP)}, pages
  19--35. IEEE, 2018.

\bibitem{shokri2017membership}
Reza Shokri, Marco Stronati, Congzheng Song, and Vitaly Shmatikov.
\newblock Membership inference attacks against machine learning models.
\newblock In {\em 2017 IEEE Symposium on Security and Privacy (SP)}, pages
  3--18. IEEE, 2017.

\bibitem{salem2018ml}
Ahmed Salem, Yang Zhang, Mathias Humbert, Pascal Berrang, Mario Fritz, and
  Michael Backes.
\newblock Ml-leaks: Model and data independent membership inference attacks and
  defenses on machine learning models.
\newblock {\em arXiv preprint arXiv:1806.01246}, 2018.

\bibitem{choo2020label}
Christopher A~Choquette Choo, Florian Tramer, Nicholas Carlini, and Nicolas
  Papernot.
\newblock Label-only membership inference attacks.
\newblock {\em arXiv preprint arXiv:2007.14321}, 2020.

\bibitem{hong2019terminalbrain}
Sanghyun Hong, Pietro Frigo, Yigitcan Kaya, Cristiano Giuffrida, and Tudor
  Dumitras.
\newblock Terminal brain damage: Exposing the graceless degradation in deep
  neural networks under hardware fault attacks.
\newblock In {\em 28th {USENIX} Security Symposium ({USENIX} Security 19)},
  pages 497--514, Santa Clara, CA, August 2019. {USENIX} Association.

\bibitem{biggio2018wild}
Battista Biggio and Fabio Roli.
\newblock Wild patterns: Ten years after the rise of adversarial machine
  learning.
\newblock {\em Pattern Recognition}, 84:317--331, 2018.

\bibitem{papernot2016towards}
Nicolas Papernot, Patrick McDaniel, Arunesh Sinha, and Michael Wellman.
\newblock Towards the science of security and privacy in machine learning.
\newblock {\em arXiv preprint arXiv:1611.03814}, 2016.

\bibitem{palmieri2015energy}
Francesco Palmieri, Sergio Ricciardi, Ugo Fiore, Massimo Ficco, and Aniello
  Castiglione.
\newblock Energy-oriented denial of service attacks: an emerging menace for
  large cloud infrastructures.
\newblock {\em The Journal of Supercomputing}, 71(5):1620--1641, 2015.

\bibitem{martin2004denial}
Thomas Martin, Michael Hsiao, Dong Ha, and Jayan Krishnaswami.
\newblock Denial-of-service attacks on battery-powered mobile computers.
\newblock In {\em Second IEEE Annual Conference on Pervasive Computing and
  Communications, 2004. Proceedings of the}, pages 309--318. IEEE, 2004.

\bibitem{krizhevsky2012imagenet}
Alex Krizhevsky, Ilya Sutskever, and Geoffrey~E Hinton.
\newblock Imagenet classification with deep convolutional neural networks.
\newblock In {\em Advances in neural information processing systems}, pages
  1097--1105, 2012.

\bibitem{jouppi2017datacenter}
Norman~P Jouppi, Cliff Young, Nishant Patil, David Patterson, Gaurav Agrawal,
  Raminder Bajwa, Sarah Bates, Suresh Bhatia, Nan Boden, Al~Borchers, et~al.
\newblock In-datacenter performance analysis of a tensor processing unit.
\newblock In {\em Proceedings of the 44th Annual International Symposium on
  Computer Architecture}, pages 1--12, 2017.

\bibitem{apple2017}
Computer Vision Machine~Learning Team.
\newblock An on-device deep neural network for face detection.
\newblock In {\em Apple Machine Learning Journal}, 2017.

\bibitem{kocher2019spectre}
Paul Kocher, Jann Horn, Anders Fogh, Daniel Genkin, Daniel Gruss, Werner Haas,
  Mike Hamburg, Moritz Lipp, Stefan Mangard, Thomas Prescher, et~al.
\newblock Spectre attacks: Exploiting speculative execution.
\newblock In {\em 2019 IEEE Symposium on Security and Privacy (SP)}, pages
  1--19. IEEE, 2019.

\bibitem{strubell2019energy}
Emma Strubell, Ananya Ganesh, and Andrew McCallum.
\newblock Energy and policy considerations for deep learning in nlp.
\newblock {\em arXiv preprint arXiv:1906.02243}, 2019.

\bibitem{bernal2012financial}
Armando Bernal, Sam Fok, and Rohit Pidaparthi.
\newblock Financial market time series prediction with recurrent neural
  networks.
\newblock {\em State College: Citeseer.[Google Scholar]}, 2012.

\bibitem{edel2016binarized}
Marcus Edel and Enrico K{\"o}ppe.
\newblock Binarized-blstm-rnn based human activity recognition.
\newblock In {\em 2016 International conference on indoor positioning and
  indoor navigation (IPIN)}, pages 1--7. IEEE, 2016.

\bibitem{cooperation2016intel}
Intel Intel.
\newblock Intel architecture instruction set extensions programming reference.
\newblock {\em Intel Corp., Mountain View, CA, USA, Tech. Rep}, pages
  319433--030, 2016.

\bibitem{markidis2018nvidia}
Stefano Markidis, Steven~Wei Der~Chien, Erwin Laure, Ivy~Bo Peng, and Jeffrey~S
  Vetter.
\newblock Nvidia tensor core programmability, performance \& precision.
\newblock In {\em 2018 IEEE International Parallel and Distributed Processing
  Symposium Workshops (IPDPSW)}, pages 522--531. IEEE, 2018.

\bibitem{hazelwood2018applied}
Kim Hazelwood, Sarah Bird, David Brooks, Soumith Chintala, Utku Diril, Dmytro
  Dzhulgakov, Mohamed Fawzy, Bill Jia, Yangqing Jia, Aditya Kalro, et~al.
\newblock Applied machine learning at facebook: A datacenter infrastructure
  perspective.
\newblock In {\em 2018 IEEE International Symposium on High Performance
  Computer Architecture (HPCA)}, pages 620--629. IEEE, 2018.

\bibitem{chung2018serving}
Eric Chung, Jeremy Fowers, Kalin Ovtcharov, Michael Papamichael, Adrian
  Caulfield, Todd Massengill, Ming Liu, Daniel Lo, Shlomi Alkalay, Michael
  Haselman, et~al.
\newblock Serving dnns in real time at datacenter scale with project brainwave.
\newblock {\em IEEE Micro}, 38(2):8--20, 2018.

\bibitem{jouppi2018motivation}
Norman Jouppi, Cliff Young, Nishant Patil, and David Patterson.
\newblock Motivation for and evaluation of the first tensor processing unit.
\newblock {\em IEEE Micro}, 38(3):10--19, 2018.

\bibitem{chen2019eyeriss}
Yu-Hsin Chen, Tien-Ju Yang, Joel Emer, and Vivienne Sze.
\newblock Eyeriss v2: A flexible accelerator for emerging deep neural networks
  on mobile devices.
\newblock {\em IEEE Journal on Emerging and Selected Topics in Circuits and
  Systems}, 9(2):292--308, 2019.

\bibitem{han2016eie}
Song Han, Xingyu Liu, Huizi Mao, Jing Pu, Ardavan Pedram, Mark~A Horowitz, and
  William~J Dally.
\newblock Eie: efficient inference engine on compressed deep neural network.
\newblock {\em ACM SIGARCH Computer Architecture News}, 44(3):243--254, 2016.

\bibitem{zhao2019automatic}
Yiren Zhao, Xitong Gao, Xuan Guo, Junyi Liu, Erwei Wang, Robert Mullins,
  Peter~YK Cheung, George Constantinides, and Cheng-Zhong Xu.
\newblock Automatic generation of multi-precision multi-arithmetic cnn
  accelerators for fpgas.
\newblock In {\em 2019 International Conference on Field-Programmable
  Technology (ICFPT)}, pages 45--53. IEEE, 2019.

\bibitem{barroso2005price}
Luiz~Andr{\'e} Barroso.
\newblock The price of performance.
\newblock {\em Queue}, 3(7):48--53, 2005.

\bibitem{li2016power}
Chao Li, Zhenhua Wang, Xiaofeng Hou, Haopeng Chen, Xiaoyao Liang, and Minyi
  Guo.
\newblock Power attack defense: Securing battery-backed data centers.
\newblock {\em ACM SIGARCH Computer Architecture News}, 44(3):493--505, 2016.

\bibitem{somani2016ddos}
Gaurav Somani, Manoj~Singh Gaur, Dheeraj Sanghi, and Mauro Conti.
\newblock Ddos attacks in cloud computing: Collateral damage to non-targets.
\newblock {\em Computer Networks}, 109:157--171, 2016.

\bibitem{xu2014power}
Zhang Xu, Haining Wang, Zichen Xu, and Xiaorui Wang.
\newblock Power attack: An increasing threat to data centers.
\newblock In {\em NDSS}, 2014.

\bibitem{xu2015measurement}
Zhang Xu, Haining Wang, and Zhenyu Wu.
\newblock A measurement study on co-residence threat inside the cloud.
\newblock In {\em 24th $\{$USENIX$\}$ Security Symposium ($\{$USENIX$\}$
  Security 15)}, pages 929--944, 2015.

\bibitem{fiore2014multimedia}
Ugo Fiore, Francesco Palmieri, Aniello Castiglione, Vincenzo Loia, and Alfredo
  De~Santis.
\newblock Multimedia-based battery drain attacks for android devices.
\newblock In {\em 2014 IEEE 11th Consumer Communications and Networking
  Conference (CCNC)}, pages 145--150. IEEE, 2014.

\bibitem{tang2017clkscrew}
Adrian Tang, Simha Sethumadhavan, and Salvatore Stolfo.
\newblock $\{$CLKSCREW$\}$: exposing the perils of security-oblivious energy
  management.
\newblock In {\em 26th $\{$USENIX$\}$ Security Symposium ($\{$USENIX$\}$
  Security 17)}, pages 1057--1074, 2017.

\bibitem{chen2009sensor}
Xiangqian Chen, Kia Makki, Kang Yen, and Niki Pissinou.
\newblock Sensor network security: a survey.
\newblock {\em IEEE Communications Surveys \& Tutorials}, 11(2):52--73, 2009.

\bibitem{anderson2003more}
Dave Anderson, Jim Dykes, and Erik Riedel.
\newblock More than an interface-scsi vs. ata.

\bibitem{efraim2014energyaware}
R.~{Efraim}, R.~{Ginosar}, C.~{Weiser}, and A.~{Mendelson}.
\newblock Energy aware race to halt: A down to earth approach for platform
  energy management.
\newblock {\em IEEE Computer Architecture Letters}, 13(1):25--28, 2014.

\bibitem{tabassi2019taxonomy}
Elham Tabassi, Kevin~J Burns, Michael Hadjimichael, Andres~D Molina-Markham,
  and Julian~T Sexton.
\newblock A taxonomy and terminology of adversarial machine learning.

\bibitem{papernot2017practical}
Nicolas Papernot, Patrick McDaniel, Ian Goodfellow, Somesh Jha, Z~Berkay Celik,
  and Ananthram Swami.
\newblock Practical black-box attacks against machine learning.
\newblock In {\em Proceedings of the 2017 ACM on Asia conference on computer
  and communications security}, pages 506--519, 2017.

\bibitem{chen2017zoo}
Pin-Yu Chen, Huan Zhang, Yash Sharma, Jinfeng Yi, and Cho-Jui Hsieh.
\newblock Zoo: Zeroth order optimization based black-box attacks to deep neural
  networks without training substitute models.
\newblock In {\em Proceedings of the 10th ACM Workshop on Artificial
  Intelligence and Security}, pages 15--26, 2017.

\bibitem{nieles2017introduction}
Michael Nieles, Kelley Dempsey, and Victoria Pillitteri.
\newblock An introduction to information security.
\newblock Technical report, National Institute of Standards and Technology,
  2017.

\bibitem{ferguson2000rfc2827}
Paul Ferguson and Daniel Senie.
\newblock rfc2827: network ingress filtering: defeating denial of service
  attacks which employ ip source address spoofing, 2000.

\bibitem{bellardo2003802}
John Bellardo and Stefan Savage.
\newblock 802.11 denial-of-service attacks: Real vulnerabilities and practical
  solutions.
\newblock In {\em USENIX security symposium}, volume~12, pages 2--2. Washington
  DC, 2003.

\bibitem{erba2019realtimeevasion}
Alessandro Erba, Riccardo Taormina, Stefano Galelli, Marcello Pogliani, Michele
  Carminati, Stefano Zanero, and Nils~Ole Tippenhauer.
\newblock Real-time evasion attacks with physical constraints on deep
  learning-based anomaly detectors in industrial control systems.
\newblock {\em CoRR}, abs/1907.07487, 2019.

\bibitem{horowitz2014computingenergy}
M.~{Horowitz}.
\newblock 1.1 computing's energy problem (and what we can do about it).
\newblock In {\em 2014 IEEE International Solid-State Circuits Conference
  Digest of Technical Papers (ISSCC)}, pages 10--14, 2014.

\bibitem{vaswani2017attention}
Ashish Vaswani, Noam Shazeer, Niki Parmar, Jakob Uszkoreit, Llion Jones,
  Aidan~N Gomez, {\L}ukasz Kaiser, and Illia Polosukhin.
\newblock Attention is all you need.
\newblock In {\em Advances in neural information processing systems}, pages
  5998--6008, 2017.

\bibitem{mikolov2013distributed}
Tomas Mikolov, Ilya Sutskever, Kai Chen, Greg~S Corrado, and Jeff Dean.
\newblock Distributed representations of words and phrases and their
  compositionality.
\newblock In {\em Advances in neural information processing systems}, pages
  3111--3119, 2013.

\bibitem{sharma2018bit}
Hardik Sharma, Jongse Park, Naveen Suda, Liangzhen Lai, Benson Chau, Vikas
  Chandra, and Hadi Esmaeilzadeh.
\newblock Bit fusion: Bit-level dynamically composable architecture for
  accelerating deep neural network.
\newblock In {\em 2018 ACM/IEEE 45th Annual International Symposium on Computer
  Architecture (ISCA)}, pages 764--775. IEEE, 2018.

\bibitem{parashar2017scnn}
Angshuman Parashar, Minsoo Rhu, Anurag Mukkara, Antonio Puglielli, Rangharajan
  Venkatesan, Brucek Khailany, Joel Emer, Stephen~W Keckler, and William~J
  Dally.
\newblock Scnn: An accelerator for compressed-sparse convolutional neural
  networks.
\newblock {\em ACM SIGARCH Computer Architecture News}, 45(2):27--40, 2017.

\bibitem{nikolic2019characterizing}
Milo{\v{s}} Nikoli{\'c}, Mostafa Mahmoud, Andreas Moshovos, Yiren Zhao, and
  Robert Mullins.
\newblock Characterizing sources of ineffectual computations in deep learning
  networks.
\newblock In {\em 2019 IEEE International Symposium on Performance Analysis of
  Systems and Software (ISPASS)}, pages 165--176. IEEE, 2019.

\bibitem{kathail2020xilinx}
Vinod Kathail.
\newblock Xilinx vitis unified software platform.
\newblock In {\em The 2020 ACM/SIGDA International Symposium on
  Field-Programmable Gate Arrays}, pages 173--174, 2020.

\bibitem{gao2018dynamic}
Xitong Gao, Yiren Zhao, {\L}ukasz Dudziak, Robert Mullins, and Cheng-zhong Xu.
\newblock Dynamic channel pruning: Feature boosting and suppression.
\newblock {\em arXiv preprint arXiv:1810.05331}, 2018.

\bibitem{hua2019channel}
Weizhe Hua, Yuan Zhou, Christopher~M De~Sa, Zhiru Zhang, and G~Edward Suh.
\newblock Channel gating neural networks.
\newblock In {\em Advances in Neural Information Processing Systems}, pages
  1886--1896, 2019.

\bibitem{xu2016automatically}
Weilin Xu, Yanjun Qi, and David Evans.
\newblock Automatically evading classifiers.
\newblock In {\em Proceedings of the 2016 network and distributed systems
  symposium}, volume~10, 2016.

\bibitem{byrd1995limited}
Richard~H Byrd, Peihuang Lu, Jorge Nocedal, and Ciyou Zhu.
\newblock A limited memory algorithm for bound constrained optimization.
\newblock {\em SIAM Journal on scientific computing}, 16(5):1190--1208, 1995.

\bibitem{hahnel2012measuring}
Marcus H{\"a}hnel, Bj{\"o}rn D{\"o}bel, Marcus V{\"o}lp, and Hermann
  H{\"a}rtig.
\newblock Measuring energy consumption for short code paths using rapl.
\newblock {\em ACM SIGMETRICS Performance Evaluation Review}, 40(3):13--17,
  2012.

\bibitem{khan2018raplinaction}
Kashif~Nizam Khan, Mikael Hirki, Tapio Niemi, Jukka~K. Nurminen, and Zhonghong
  Ou.
\newblock Rapl in action: Experiences in using rapl for power measurements.
\newblock {\em ACM Trans. Model. Perform. Eval. Comput. Syst.}, 3(2), March
  2018.

\bibitem{sen2018gpumeasure}
S.~{Sen}, N.~{Imam}, and C.~{Hsu}.
\newblock Quality assessment of gpu power profiling mechanisms.
\newblock In {\em 2018 IEEE International Parallel and Distributed Processing
  Symposium Workshops (IPDPSW)}, pages 702--711, 2018.

\bibitem{ott2019fairseq}
Myle Ott, Sergey Edunov, Alexei Baevski, Angela Fan, Sam Gross, Nathan Ng,
  David Grangier, and Michael Auli.
\newblock fairseq: A fast, extensible toolkit for sequence modeling.
\newblock In {\em Proceedings of NAACL-HLT 2019: Demonstrations}, 2019.

\bibitem{liu2019roberta}
Yinhan Liu, Myle Ott, Naman Goyal, Jingfei Du, Mandar Joshi, Danqi Chen, Omer
  Levy, Mike Lewis, Luke Zettlemoyer, and Veselin Stoyanov.
\newblock Roberta: A robustly optimized bert pretraining approach.
\newblock {\em arXiv preprint arXiv:1907.11692}, 2019.

\bibitem{devlin2018bert}
Jacob Devlin, Ming-Wei Chang, Kenton Lee, and Kristina Toutanova.
\newblock Bert: Pre-training of deep bidirectional transformers for language
  understanding.
\newblock {\em arXiv preprint arXiv:1810.04805}, 2018.

\bibitem{wang2018glue}
Alex Wang, Amanpreet Singh, Julian Michael, Felix Hill, Omer Levy, and Samuel~R
  Bowman.
\newblock Glue: A multi-task benchmark and analysis platform for natural
  language understanding.
\newblock {\em arXiv preprint arXiv:1804.07461}, 2018.

\bibitem{wang2019superglue}
Alex Wang, Yada Pruksachatkun, Nikita Nangia, Amanpreet Singh, Julian Michael,
  Felix Hill, Omer Levy, and Samuel Bowman.
\newblock Superglue: A stickier benchmark for general-purpose language
  understanding systems.
\newblock In {\em Advances in Neural Information Processing Systems}, pages
  3261--3275, 2019.

\bibitem{ott2018scaling}
Myle Ott, Sergey Edunov, David Grangier, and Michael Auli.
\newblock Scaling neural machine translation.
\newblock {\em arXiv preprint arXiv:1806.00187}, 2018.

\bibitem{edunov2018understanding}
Sergey Edunov, Myle Ott, Michael Auli, and David Grangier.
\newblock Understanding back-translation at scale.
\newblock {\em arXiv preprint arXiv:1808.09381}, 2018.

\bibitem{ng2019WMT}
Nathan Ng, Kyra Yee, Alexei Baevski, Myle Ott, Michael Auli, and Sergey Edunov.
\newblock Facebook fair's {WMT19} news translation task submission.
\newblock {\em CoRR}, abs/1907.06616, 2019.

\bibitem{koehn2007moses}
Philipp Koehn, Hieu Hoang, Alexandra Birch, Chris Callison-Burch, Marcello
  Federico, Nicola Bertoldi, Brooke Cowan, Wade Shen, Christine Moran, Richard
  Zens, et~al.
\newblock Moses: Open source toolkit for statistical machine translation.
\newblock In {\em Proceedings of the 45th annual meeting of the association for
  computational linguistics companion volume proceedings of the demo and poster
  sessions}, pages 177--180, 2007.

\bibitem{sennrich2015subword}
Rico Sennrich, Barry Haddow, and Alexandra Birch.
\newblock Neural machine translation of rare words with subword units.
\newblock {\em CoRR}, abs/1508.07909, 2015.

\bibitem{ng2019facebook}
Nathan Ng, Kyra Yee, Alexei Baevski, Myle Ott, Michael Auli, and Sergey Edunov.
\newblock Facebook fair's wmt19 news translation task submission.
\newblock {\em arXiv preprint arXiv:1907.06616}, 2019.

\bibitem{akamai_ddos_report}
Ryan Barnett.
\newblock The dark side of apis: Denial of service attacks.

\bibitem{azure_limits}
Microsoft Azure.
\newblock Request limits for translator.

\bibitem{he2016deep}
Kaiming He, Xiangyu Zhang, Shaoqing Ren, and Jian Sun.
\newblock Deep residual learning for image recognition.
\newblock In {\em Proceedings of the IEEE conference on computer vision and
  pattern recognition}, pages 770--778, 2016.

\bibitem{huang2017densely}
Gao Huang, Zhuang Liu, Laurens Van Der~Maaten, and Kilian~Q Weinberger.
\newblock Densely connected convolutional networks.
\newblock In {\em Proceedings of the IEEE conference on computer vision and
  pattern recognition}, pages 4700--4708, 2017.

\bibitem{sandler2018mobilenetv2}
Mark Sandler, Andrew Howard, Menglong Zhu, Andrey Zhmoginov, and Liang-Chieh
  Chen.
\newblock Mobilenetv2: Inverted residuals and linear bottlenecks.
\newblock In {\em Proceedings of the IEEE conference on computer vision and
  pattern recognition}, pages 4510--4520, 2018.

\bibitem{kim2017novel}
Dongyoung Kim, Junwhan Ahn, and Sungjoo Yoo.
\newblock A novel zero weight/activation-aware hardware architecture of
  convolutional neural network.
\newblock In {\em Design, Automation \& Test in Europe Conference \& Exhibition
  (DATE), 2017}, pages 1462--1467. IEEE, 2017.

\bibitem{gowal2018effectiveness}
Sven Gowal, Krishnamurthy Dvijotham, Robert Stanforth, Rudy Bunel, Chongli Qin,
  Jonathan Uesato, Relja Arandjelovic, Timothy Mann, and Pushmeet Kohli.
\newblock On the effectiveness of interval bound propagation for training
  verifiably robust models.
\newblock {\em arXiv preprint arXiv:1810.12715}, 2018.

\bibitem{goodfellow2014explaining}
Ian~J Goodfellow, Jonathon Shlens, and Christian Szegedy.
\newblock Explaining and harnessing adversarial examples.
\newblock {\em International Conference on Learning Representations (ICLR)},
  2015.

\bibitem{wong2020fast}
Eric Wong, Leslie Rice, and J.~Zico Kolter.
\newblock Fast is better than free: Revisiting adversarial training, 2020.

\bibitem{chen2020stateful}
Steven Chen, Nicholas Carlini, and David Wagner.
\newblock Stateful detection of black-box adversarial attacks.
\newblock In {\em Proceedings of the 1st ACM Workshop on Security and Privacy
  on Artificial Intelligence}, SPAI '20, page 30–39, New York, NY, USA, 2020.
  Association for Computing Machinery.

\bibitem{shumailov2020ctt}
Ilia Shumailov, Yiren Zhao, Robert Mullins, and Ross Anderson.
\newblock Towards certifiable adversarial sample detection.
\newblock In {\em Proceedings of the 13th ACM Workshop on Artificial
  Intelligence and Security}, AISec'20, page 13–24, New York, NY, USA, 2020.
  Association for Computing Machinery.

\bibitem{henderson2020towards}
Peter Henderson, Jieru Hu, Joshua Romoff, Emma Brunskill, Dan Jurafsky, and
  Joelle Pineau.
\newblock Towards the systematic reporting of the energy and carbon footprints
  of machine learning.
\newblock {\em arXiv preprint arXiv:2002.05651}, 2020.

\bibitem{lacoste2019quantifying}
Alexandre Lacoste, Alexandra Luccioni, Victor Schmidt, and Thomas Dandres.
\newblock Quantifying the carbon emissions of machine learning.
\newblock {\em arXiv preprint arXiv:1910.09700}, 2019.

\bibitem{goel2012techniques}
Bhavishya Goel, Sally~A McKee, and Magnus Sj{\"a}lander.
\newblock Techniques to measure, model, and manage power.
\newblock In {\em Advances in Computers}, volume~87, pages 7--54. Elsevier,
  2012.

\bibitem{garcia2019estimation}
Eva Garc{\'\i}a-Mart{\'\i}n, Crefeda~Faviola Rodrigues, Graham Riley, and
  H{\aa}kan Grahn.
\newblock Estimation of energy consumption in machine learning.
\newblock {\em Journal of Parallel and Distributed Computing}, 134:75--88,
  2019.

\bibitem{butts2000static}
J~Adam Butts and Gurindar~S Sohi.
\newblock A static power model for architects.
\newblock In {\em Proceedings 33rd Annual IEEE/ACM International Symposium on
  Microarchitecture. MICRO-33 2000}, pages 191--201. IEEE, 2000.

\end{thebibliography}

\appendix

\begin{figure}[]
    \centering
    \includegraphics[width=\linewidth]{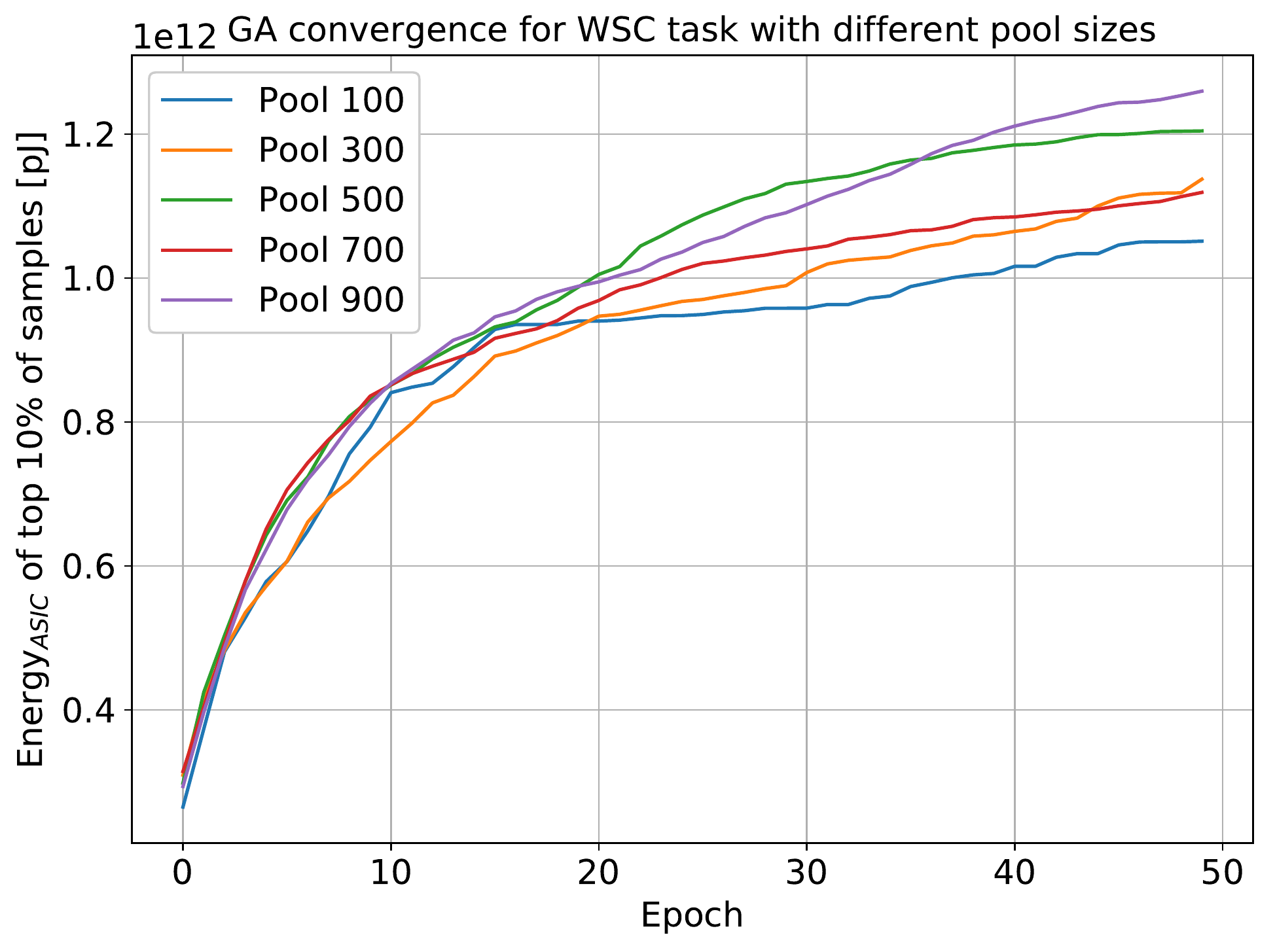}
    \caption{GA performance with WSC task from GLUE Benchmark running on GPUs. Words of size 29 are evaluated with pool sizes of 100, 300, 500, 700 and 900.}
    \label{fig:ga_training}
\end{figure}

\begin{algorithm}[]
\SetAlgoLined
\KwResult{S}
\nonl\textcolor{blue}{\textit{initialise a random pool of inputs}};\\
$S = \{S_0, S_1, ..., S_n\}$\;
\While{$i < K $}{
    \nonl \textcolor{blue}{\textit{Profile the inputs to get fitness scores};\ \nonl $\Rightarrow$ \textbf{latency} or \textbf{energy}}\\
    $P$ = \textsf{Fitness}($S$);\\
    \nonl \textcolor{blue}{\textit{Pick top performing samples}};\\
    $\hat{S}$ = \textsf{Select}($P$, $S$);\\

    \If{NLP}{
        $S$ = \textsf{MutateNLP}($\hat{S}$);\\
        \nonl \textcolor{blue}{Concatenate samples A, B};\\
        \nonl \textcolor{blue}{$\Rightarrow S=\textsf{LeftHalf}(A) + \textsf{RightHalf}(B)$};\\
        \nonl \textcolor{blue}{$\Rightarrow S=\textsf{RandomlyMutate}(S)$};\\
    }
    
    \If{CV}{
        $S$ = \textsf{MutateCV}($\hat{S}$);\\
        \nonl \textcolor{blue}{Concatenate samples A, B, and a random mask};\\
        \nonl \textcolor{blue}{$\Rightarrow A*mask + (1-mask)*B$};\\
    }
}\;
\caption{Sponge samples through a Genetic Algorithm}
\label{alg:ga}
\end{algorithm}

\newpage

\section{Parameter Choices}
\label{apdx:sec:parameter_choices}

We have thoroughly evaluated different parameter choices for the sponge attack and found that a small pool size and a relatively short number of GA iterations should be sufficient for a large number of tasks.

\Cref{fig:ga_training} shows the performance of sponge samples on the RoBERTa model for the Winograd Schema Challenge (WSC) with different pool sizes and varying input sequence length.
The horizontal axis shows the number of GA iterations.
In terms of pool size of the GA, although there is an increase in performance for larger pool sizes, the increase is marginal.
Also, smaller pool sizes significantly reduce the runtime for the attack.
From the hardware perspective, using a large pool size might trigger GPUs to throttle, so that the runtime will be further increased.
We observed that the convergence is consistently faster for smaller input sequences.
This is mainly because the complexity of the search is less.
In practice, we found almost all input sequence lengths we tested plateau within 100 GA iterations; even going to over 1000 iterations gives only a small increase in performance.
For these reasons, for the experiments presented below, we report the results of the attack with a pool size of 1000 for GLUE and Computer Vision benchmarks and 1000 for translation tasks.
We use 1000 GA iterations for all benchmarks tested.
When displaying results, we normally use sponge for sponge examples produced using the GA and use sponge L-BFGS to identify sponge examples generated using L-BFGS.

\section{Energy Cost Factors}
\label{apdx:sec:energy}

Energy cost is a combination of static and dynamic energy.

$$E = (P_{\mathsf{static}} + P_{\mathsf{dynamic}}) \times t$$

Static power refers to the consumption of the circuitry in an idle state~\cite{goel2012techniques} there are multiple models to estimate this depending on the technology~\cite{goel2012techniques,garcia2019estimation,butts2000static}.
In this paper, we follow a coarse-grained approach. Cycle-accurate hardware simulation incurs a large run-time, but a coarse-grained energy simulator provides enough resolution to indicate the energy-consuming samples while using significantly less time per round of simulation.

\begin{equation}
\label{eqn:static}
P_{\mathsf{static}} =
\sum{I_{\mathsf{leakage}}\times V_\mathsf{{core}}}
= \sum{I_{s} \times (e^{\frac{qV_d}{kT}}-1})\times V_{\mathsf{core}}
\end{equation}

where $I_{s}$ is the reverse saturation current;
$V_{d}$ is the diode voltage;
$k$ is Boltzmann's constant;
$q$ is the electronic charge;
$T$ is temperature and $V_{\mathsf{core}}$ is the supply voltage.

Dynamic power refers to consumption from charging and discharging the circuitry~\cite{garcia2019estimation}.

\begin{equation}
\label{eqn:dynamic}
P_{\mathsf{dynamic}} = \alpha \times C \times V^{2}_{core} \times f
\end{equation}

Here, $\alpha$ refers to the activity factor \ie~components that are currently consuming power; $V_{d}$ is the source voltage; $C$ is the capacitance; $f$ is the clock frequency. Ultimately an attacker attempts to solve an optimisation problem

\begin{equation}
\begin{aligned}
\max_{} E, \textrm{ where } E =
\big{(}[\sum{I_{s} \times (\overbrace{e^{\frac{qV_d}{kT}-1}}}^{\mathclap{\substack{%
\text{\color{blue}overheat or increase overall consumption}\\~\\~
}}})\times V_{\mathsf{core}}]\\ + [\underbrace{\textstyle \alpha}_{\mathclap{\text{\color{blue} more activity of the board}}} \times C \times V^{2}_{core} \times \overbrace{f}^{\mathclap{\text{\color{blue}throttle or exploit load predictor}}}]\big{)} \times \underbrace{t}_{\mathclap{\text{\color{blue}run for longer or exploit the predictor}}}.
\end{aligned}
\end{equation}

For all parameters considered in the equation,
only four can be manipulated by the adversary described in~\Cref{sec:adversary}: $T$, $\alpha$, $f$ and $t$.
Of these, frequency and temperature cannot be controlled directly, but are affected through optimisations performed by the computing hardware.
As we assume a single GPU, CPU or ASIC, we focus on the activity ratio $\alpha$, the time $t$ and the switching power from flipping the state of transistors.
The execution time $t$ and activity ratio $\alpha$ link tightly to the number of operations and memory accesses performed.
In the temporal dimension, attackers might trigger unnecessary passes of a compute-intensive block; in the spatial domain, attackers can turn sparse operations into dense ones.
These temporal and spatial attack opportunities can significantly increase the number of memory and arithmetic operations and thus create an increase in $\alpha$ and $t$ to maximise energy usage.

\section{Domain Specific Optimisations}
\label{apdx:sec:reasons}
In~\Cref{sec:eee_algo} we outlined the genetic algorithm we used to find sponge samples. That approach is generic. Here we describe how we can improve the effectiveness of the genetic algorithm through domain-specific optimisations. 

First, for NLP tasks, the greatest impact on performance was acquired from exploiting the encoding schemes used.
While the genetic algorithm was fast to pick up this vulnerability, it struggled with efficiency around the mid-point, where the parents were concatenated. For example, when trying to break down individual sub-words to more tokens, we observed the GA inserting backslashes into the samples. When concatenated, we saw cases where two non-backslashes followed each other, meaning the GA was losing on a couple of characters. As a solution, we probabilistically flipped the halves and saw a slight improvement.

For CV tasks, we observed that random samples were always classified as belonging to the same class. Furthermore, random samples had very low internal density. 
We hypothesize that this has to do with the fact that on random samples there are very few class features, as opposed to what is observed in natural samples. 
As the GA improvement largely depends on randomness, that meant that we often observed that after merging two highly dense parents, uniform randomness across all pixels was decreasing sparsity to the level of random samples. 
In other words, uniform randomness was diluting class features. To counter this phenomenon, instead of applying uniform randomness across all pixel values, we resorted to diluting only 1\% of them. 
That led to a bigger improvement in the whole population pool. Furthermore, after observing that the density is class-dependent, it became apparent that to preserve diversity in the pool it was important to keep samples from multiple classes. For this, we tried to ensure that at least 20 different classes were preserved in the pool. 

We attempted to use domain knowledge and tried adding operations like rotation, transposition and re-scaling into the mutation process, yet we found that these did not lead to significant improvements.

\section{Understanding Sponges and Their Performance}
\label{apdx:sec:understanding}

\begin{figure*}[h]
    \centering
    \subfloat{{\includegraphics[width=0.45\textwidth]{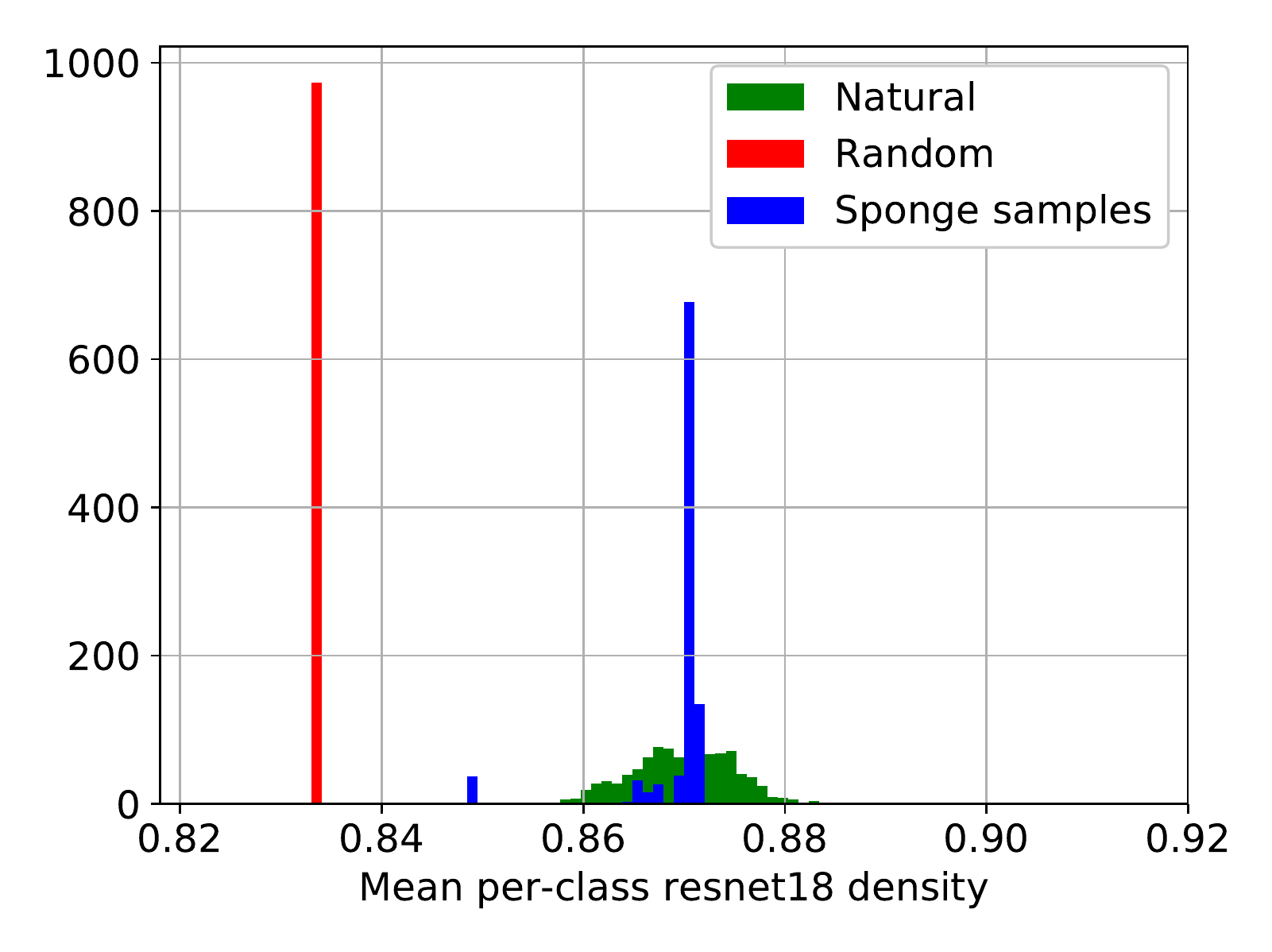} }}%
    % \qquad
    \subfloat{{\includegraphics[width=0.45\textwidth]{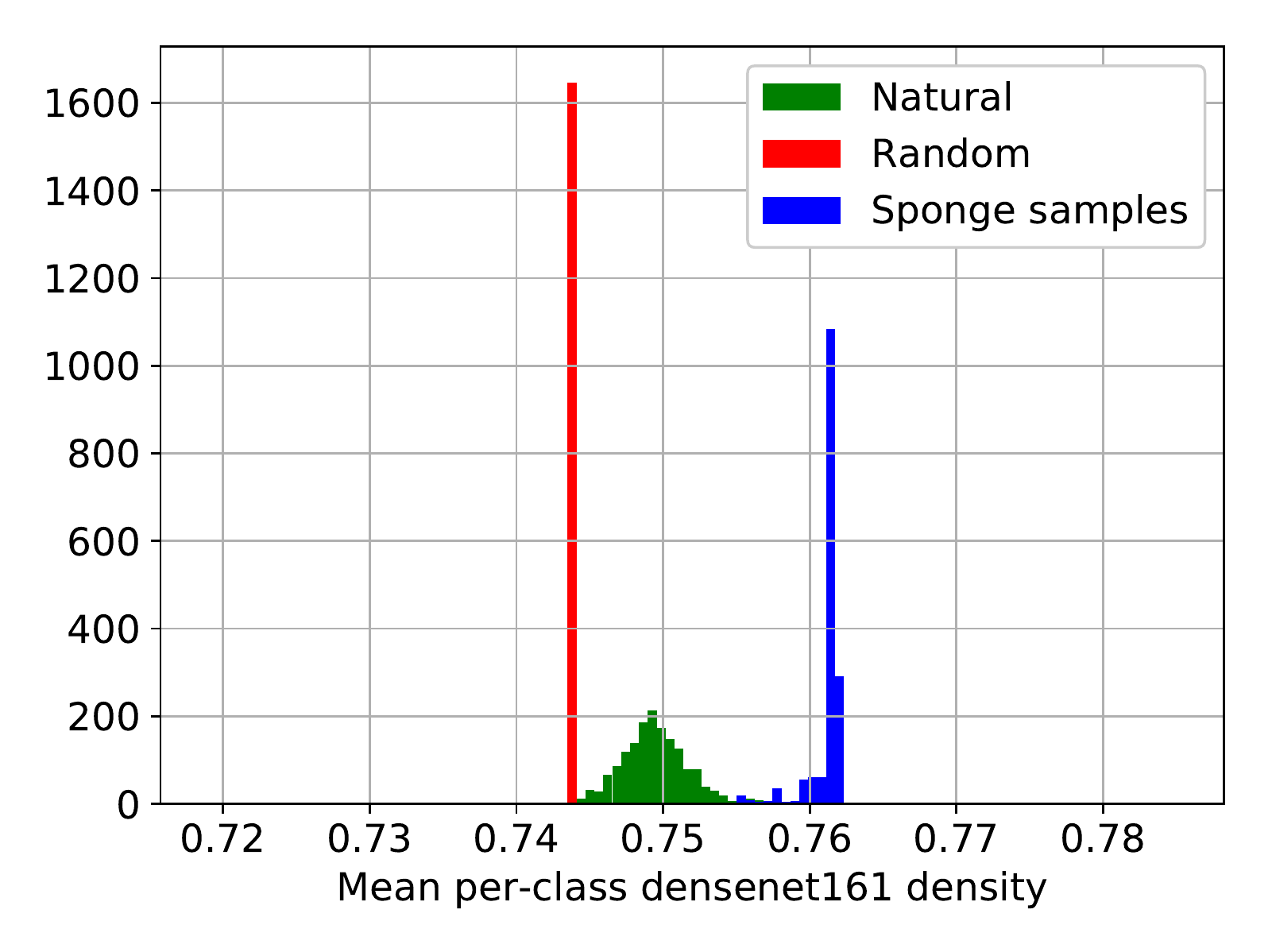}}}%
    \caption{
        Per-class mean density of samples evaluated
        on ResNet18 and DenseNet161.
        The natural samples are from the validation set
        and are compared to 50 000 randomly generated samples and 1000 Sponge GA samples.
        The scales are normalised to form a probability density.}
    \label{fig:perclassconsumption}
\end{figure*}

To better understand the results, we present \Cref{fig:perclassconsumption} which shows per-class density distributions of natural, random and sponge samples.
There are 50,000 random and natural samples respectively and 1,000 sponge samples, with the bars normalised to form a probability density.

The first thing that becomes apparent is that randomly generated samples on CV models cost significantly less energy because many activations are not on.
On average, random samples result in a sparser computation -- around $4\%$ more sparse for ResNet18 -- and our simulator costs for natural samples are around $4-7$\% higher than the costs of random samples.
Second, a surprising finding is that the most and least sparse samples are clustered in a handful of classes.
In other words, certain classes have inputs that are more expensive than others in terms of energy.
For ResNet-18, the most sparse classes are `wing' and `spotlight' while the least sparse are `greenhouse' and `howler monkey'.
We observe a similar effect for larger ResNet variants and also DenseNets, although the energy gap is smaller on DenseNets.
Interestingly, we see that energy-expensive classes are consistent across different architectures, and we further demonstrate this class-wise transferability in \Cref{apdx:sec:class_transferability}.
Ultimately, this phenomenon implies that it is possible to burn energy or slow a system without much preparation, by simply bombarding the model with natural samples from energy-consuming classes.
Finally, we see that the sponge samples are improving the population performance and tend to outperform natural sample. We observe that it is easier for sponge to outperform all natural samples for DensNets of different size, yet it struggles to outperform all of the ResNets. We further measure the energy performance statistically in~\Cref{apdx:sec:measuring_cv}.

\begin{figure*}[h]
    \centering
    \subfloat[][Energy vs Power]{{\includegraphics[width=0.45\linewidth]{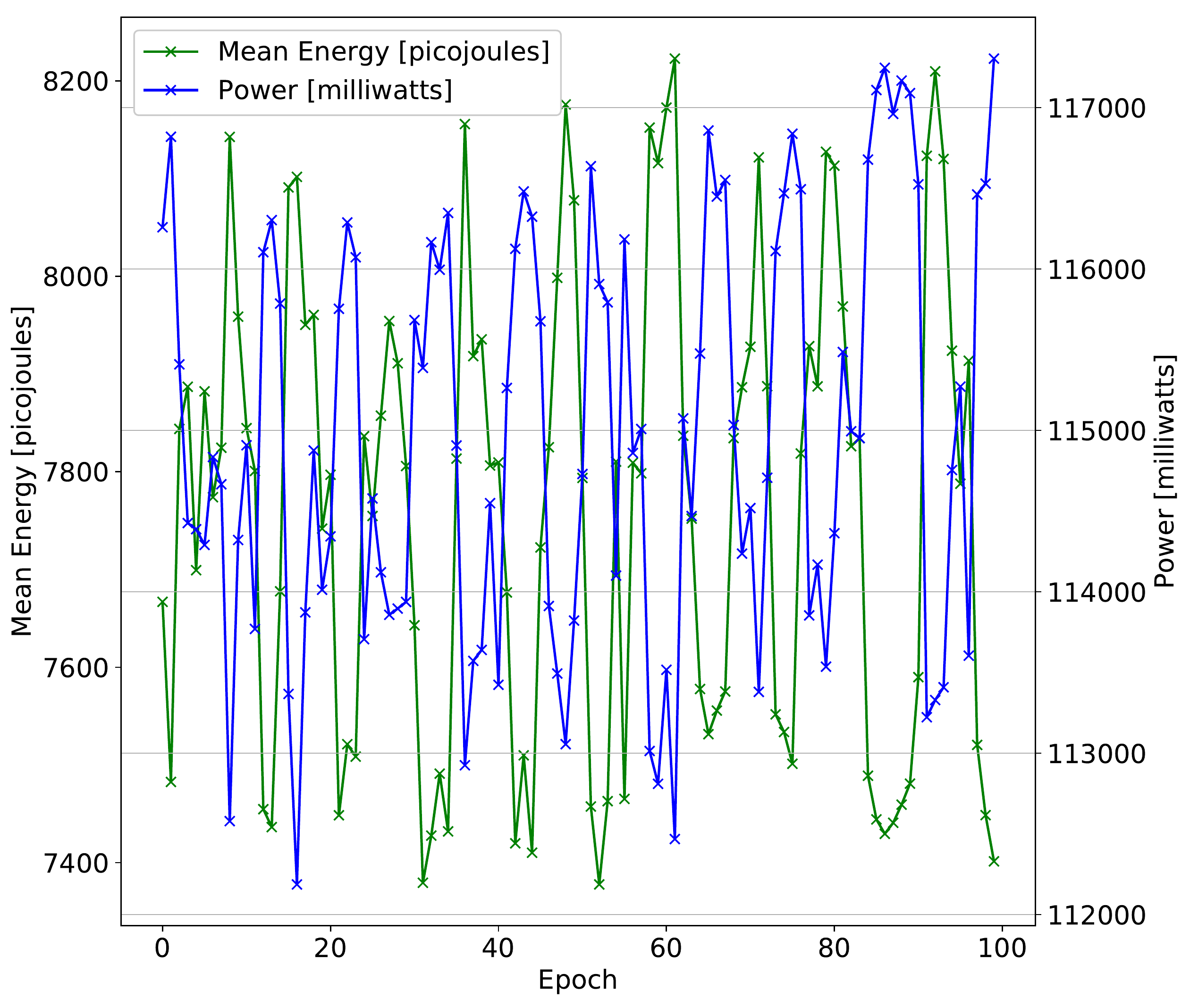}}}
    %
    % \qquad
    \subfloat[][Energy vs Time]{{\includegraphics[width=0.45\linewidth]{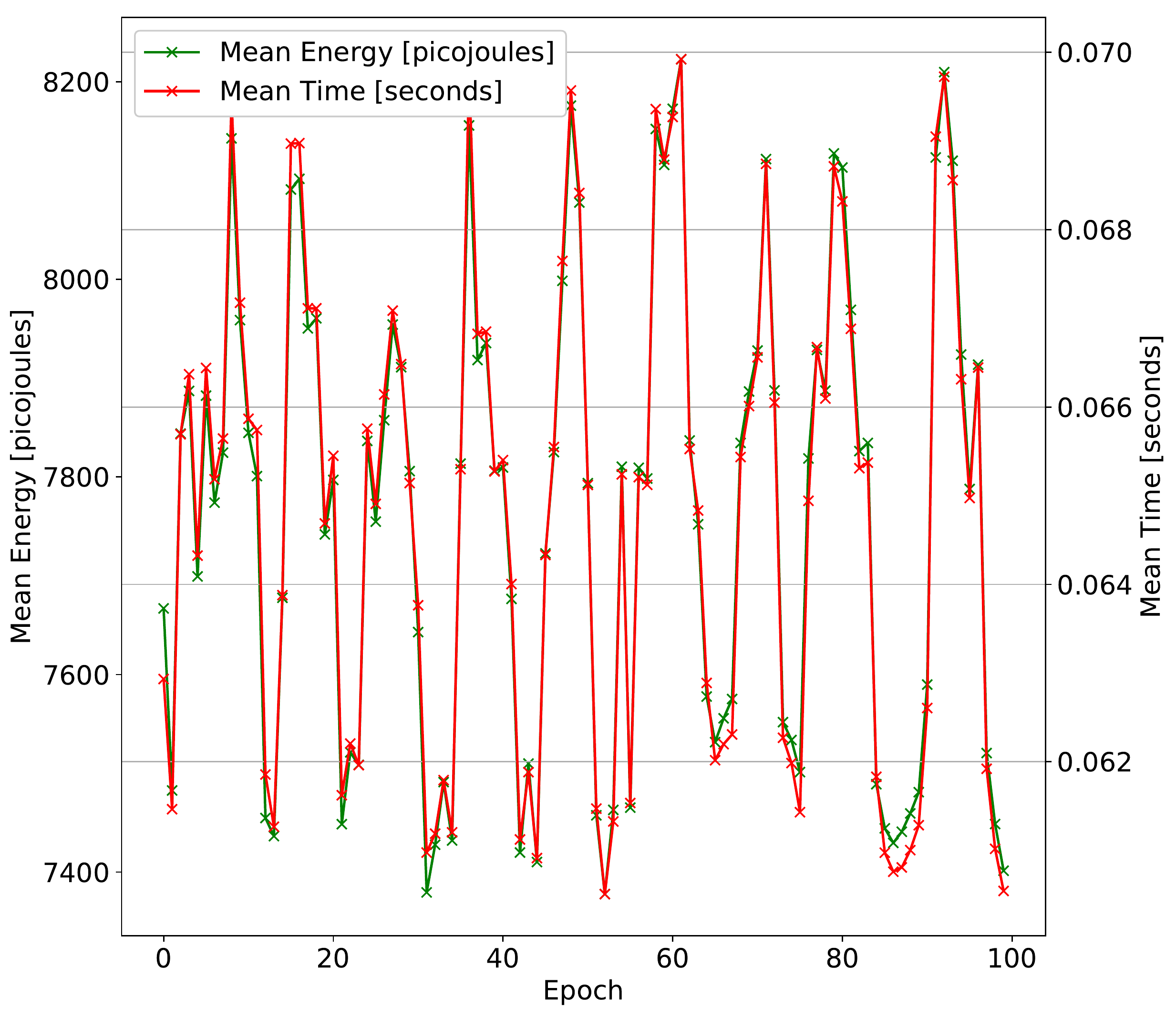}}}
    \caption{ResNet-18 solving ImageNet-2017 without any rate limiting with increasing internal density.}
    \label{fig:eng_to_power}
\end{figure*}

\begin{figure*}[h]
    \centering
    \subfloat[][First 5000 samples discarded]{{\includegraphics[width=0.45\linewidth]{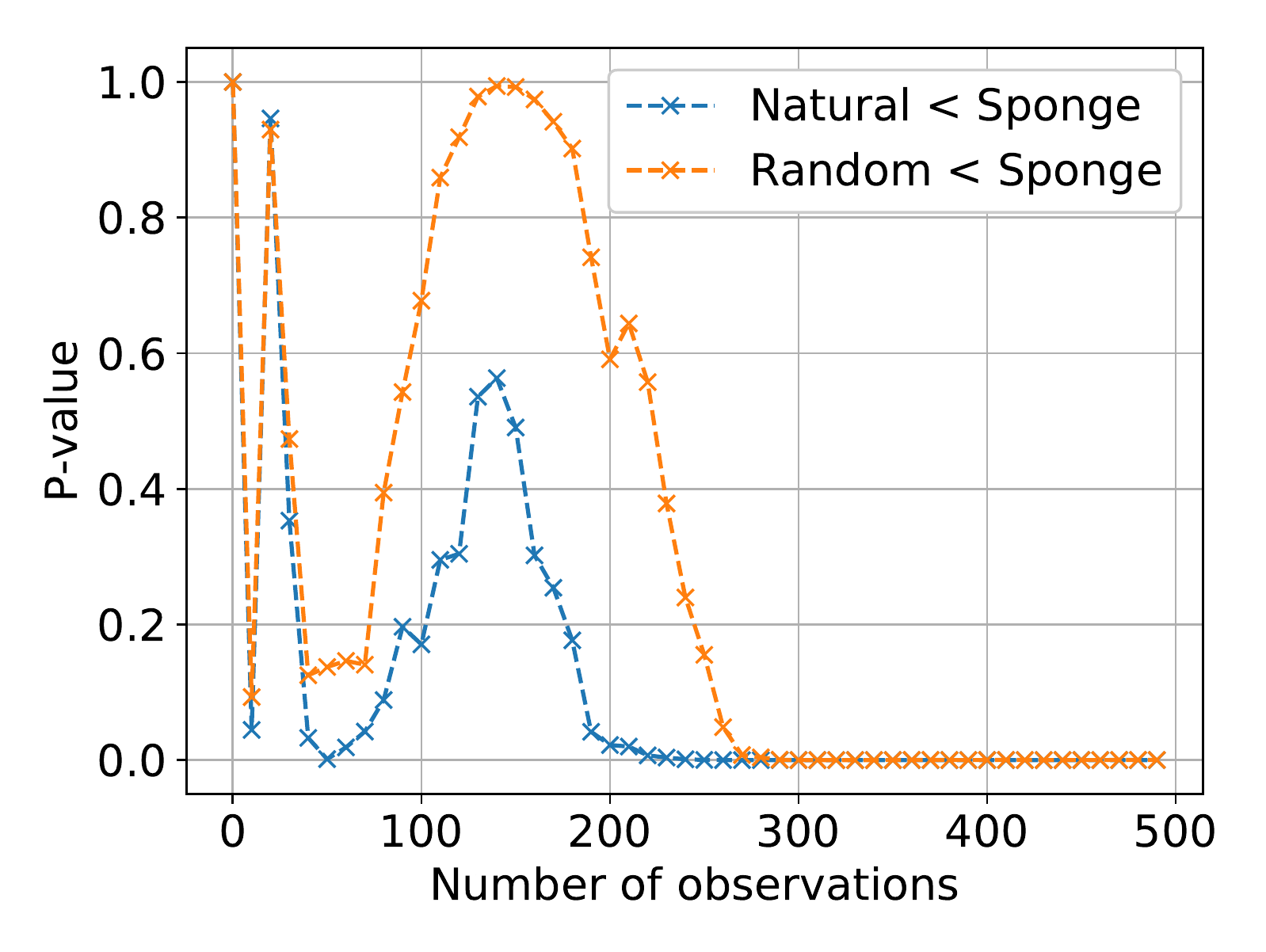}}}
    %
    % \qquad
    \subfloat[][First 30000 samples discarded]{{\includegraphics[width=0.45\linewidth]{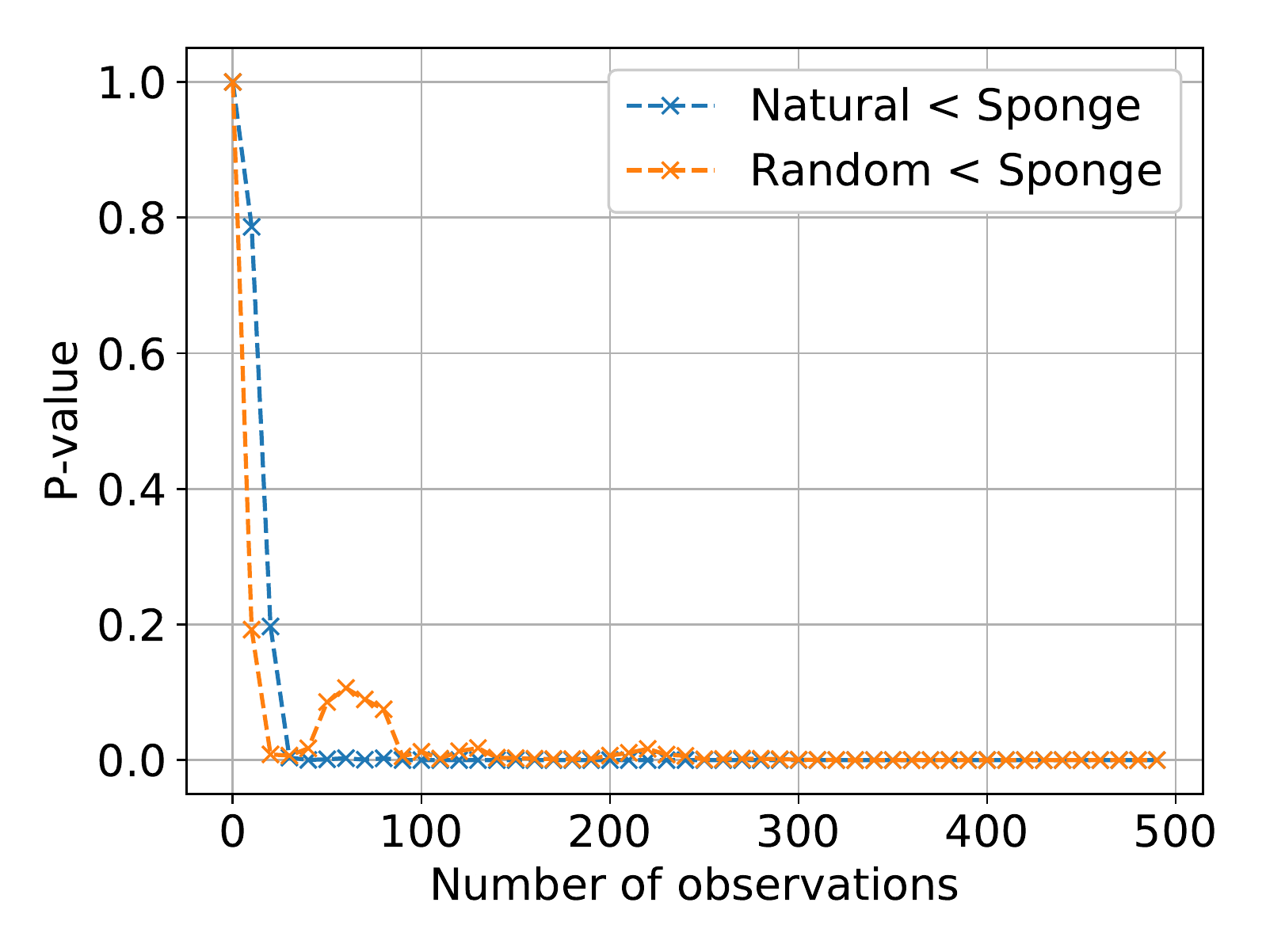}}}
    \caption{Mann-Whitney test on CPU measured Mobilenet execution. Number of observations is shown on x-axis and p-value on the y-axis.}
    \label{fig:p_value}
\end{figure*}

\subsection{Measuring Difficulties and Statistical Analysis}
\label{apdx:sec:measuring_cv}

Although we have presented in~\Cref{sec:eval:cv} that sponge attacks cause ASIC energy consumption to rise for computer vision tasks, it is still unclear what this translates to real life.

If one were to directly measure the CPU or GPU load per adversarial sample, interpreting it would be hard, especially when one talks about the energy cost improvements in the order of around 5\% for ResNet18 and 3\% as for DenseNet101.
As is mentioned in~\Cref{sec:methodology:definitions} the main energy costs include the frequency of switching activities, voltage and clock frequency.
Due to the heat impact from voltage and clock frequency, a large number of different optimisations are deployed by the hardware.
Here, the optimisations try to balance multiple objectives -- they try to be as performant as they can, whilst being as energy efficient as possible and also maintain reliability.
Modern CPUs and GPUs have several performance modes
between which the hardware can switch.
For example, official Nvidia documentation lists 15 different performance modes.
% https://docs.nvidia.com/deploy/nvml-api/group__nvmlDeviceEnumvs.html

\Cref{fig:eng_to_power} shows measurements taken during the sponge GA attack running against ResNet-18.
The x-axis shows the number of epochs, with each epoch the internal density is increasing from 0.75\% to 0.8\%.
In (a), the right y-axis shows mean energy readings per sample, whereas left y-axis shows mean power readings per-sample.
In (b) the left y-axis shows mean latency values per-sample.

The amount of power consumed is strongly correlated to the amount of time taken by each sample.
When the GPU speeds up, it consumes more energy but requires less time, but the rise in temperature causes the hardware then to go to a more conservative mode to cool down.
We observe this heating and cooling cycle with all tasks running on GPUs, making it hard to measure the absolute performance and the attack impact. We can however measure the performance statistically.
First, we turn to a question of

\begin{center}
\textit{Can we detect energy differences between Natural, Random and Sponge samples?}
\end{center}

To investigate the relationship between the samples we use Mann-Whitney-Wilcoxon U test (U-test), a nonparametric test for the difference between distributions. With three classes of samples, we need three pairwise comparisons. For each one, the null hypothesis that the distributions of energy consumed by the samples are identical. The complement hypothesis is that of a difference between distributions. 

The U-test is based on three main assumptions:
\begin{itemize}
    \item Independence between samples;
    \item The dependent variable is at least ordinal;
    \item The samples are random.
\end{itemize}

The first assumption is fulfilled since no sample belongs to more than one category i.e. natural, random and sponge. The second assumption is satisfied by the fact that both time and energy are cardinal variables. The third assumption, however, is harder to satisfy. 

The cause of this lies in the closed nature of hardware optimisations: although some of the techniques are known, the exact parameters are unknown. Furthermore, it is hard to achieve the same state of the hardware even through power cycling. As was mentioned in~\Cref{sec:methodology:definitions} temperature affects energy directly, and it is hard to make sure that the hardware always comes back to the same state. 

To minimise temperature effects we apply the load of natural, attack and random samples and wait until the temperature stabilises. That takes approximately 30000 samples. The order of the samples is random, and at this point, it can be assumed that all of the data and instruction caches are filled. Finally, because the samples are randomly shuffled, all of the predictive optimisations will work with the same probability for each of the classes. 

For these reasons, we believe it is safe to assume that the samples themselves are random in that the effect of hardware optimisations is random so that the last assumption of the Mann-Whitney test is fulfilled.

Using this test we can do a pairwise comparison of the natural, random and sponge samples. The test indicates that the three types of samples generate energy consumption distributions that are statistically different (one-sided test, p-value=0.000) for mobilenet executed on a CPU. On a practical level, the amount of energy consumed by sponge samples is 1.5\% higher on a CPU and >7\% on ASIC. We could not evaluate the energy recordings on a GPU, as the standard deviation was over 15\% which becomes worse as temperature increases. \Cref{fig:p_value} shows the confidence of the Mann-Whitney test with mobilenet measured on the CPU as a function of the number of observations. The number of observations is on the x-axis, and the p-value on the y-axis. As can be seen, in a stable environment i.e. the temperature has stabilised, after about 100 observations per class, the differences become statistically significant at any reasonable confidence level. A similar trend is observed for unstable temperature environment, but around three times more data is required. That means that in practice, about 100--300 observations per class are sufficient to differentiate between classes with high confidence.

% \section{Examples of Sponges}
% \input{sections/images/sponges/sponges_ex}

\section{Energy--Time Relationship of Sponge Examples}
\begin{table*}[!h]
\centering
\begin{adjustbox}{scale=0.8,center}
\begin{tabular}{@{}lcccccc@{}}
% \begin{tabularx}{\linewidth}{lcc *{6}{Y} }
\toprule
& 
& \multicolumn{2}{c}{GPU Energy [mJ]}
& \multicolumn{2}{c}{GPU Time [mS]}
& 
\\
& Input size 
& Natural
& Sponge
& Natural
& Sponge 
& Time $\times$ / Energy $\times$
\\
\midrule
\multicolumn{7}{l}{\textit{\underline{SuperGLUE Benchmark with \cite{liu2019roberta}}}} \vspace{3mm}\\
\multirow{3}{*}{CoLA} & 15 & 
$1.00\times$ & \multicolumn{1}{c|}{$\mathbf{1.11}\times$} 
& $\mathbf{1.00}\times$ & \multicolumn{1}{c|}{$0.92\times$}
& $1.21$ 
\\
& 30
& $1.00\times$  & \multicolumn{1}{c|}{$\mathbf{1.28}\times$} 
& $\mathbf{1.00}\times$ & \multicolumn{1}{c|}{$0.82\times$} 
& $1.56$
\\
& 50
& $1.00\times$ & \multicolumn{1}{c|}{$\mathbf{2.06}\times$} 
& $1.00\times$ & \multicolumn{1}{c|}{$\mathbf{1.27}\times$} 
& $1.62$
\\ \\

\multirow{3}{*}{MNLI}
& 15
& $1.00\times$  & \multicolumn{1}{c|}{$\mathbf{1.12}\times$} 
& $1.00\times$  &  \multicolumn{1}{c|}{$0.95\times$} 
& $1.26$ 
\\
& 30 
& $1.00\times$  & \multicolumn{1}{c|}{$\mathbf{1.51}\times$} 
& $1.00\times$  & \multicolumn{1}{c|}{$1.03\times$} 
& $1.46$ 
\\
& 50 
& $1.00\times$  & \multicolumn{1}{c|}{$\mathbf{2.16}\times$} 
& $1.00\times$ & \multicolumn{1}{c|}{$\mathbf{1.30}\times$} 
& $1.66$
\\ \\

\multirow{3}{*}{WSC}
& 15 
& $1.00\times$  & \multicolumn{1}{c|}{$\mathbf{8.89}\times$} 
& $1.00\times$  & \multicolumn{1}{c|}{$\mathbf{5.51}\times$} 
& $1.61$ 
\\
& 30 
& $1.00\times$  & \multicolumn{1}{c|}{$\mathbf{16.13}\times$} 
& $1.00\times$  & \multicolumn{1}{c|}{$\mathbf{11.04}\times$} 
& $1.46$ 
\\
& 50 
& $1.00\times$  & \multicolumn{1}{c|}{$\mathbf{26.64}\times$} 
& $1.00\times$ & \multicolumn{1}{c|}{$\mathbf{20.56}\times$}  
& $1.29$ 
\\ \\
\midrule

\multicolumn{7}{l}{\textit{\underline{WMT14/16 with \cite{ott2018scaling}}}}\vspace{3mm}\\
\multirow{1}{*}{En$\rightarrow$Fr} 
& 15
& $1.00\times$ & \multicolumn{1}{c|}{$\mathbf{4.32}\times$} 
& $1.00\times$ & \multicolumn{1}{c|}{$\mathbf{3.89}\times$} 
& $1.11$ 
\\ 

\multirow{1}{*}{En$\rightarrow$De} 
& 15
& $1.00\times$ & \multicolumn{1}{c|}{$\mathbf{27.84}\times$} 
& $1.00\times$ & \multicolumn{1}{c|}{$\mathbf{24.18}\times$} 
& $1.15$ 
\\ \\
\midrule

\multicolumn{7}{l}{\textit{\underline{WMT18 with \cite{edunov2018understanding}}}}\vspace{3mm}\\

\multirow{1}{*}{En$\rightarrow$De} 
& 15
& $1.00\times$ & \multicolumn{1}{c|}{$\mathbf{30.81}\times$} 
& $1.00\times$ & \multicolumn{1}{c|}{$\mathbf{26.49}\times$} 
& $1.16$ 
\\ \\
\midrule

\multicolumn{7}{l}{\textit{\underline{WMT19 with \cite{ng2019facebook}}}}\vspace{3mm}\\
\multirow{1}{*}{En$\rightarrow$Ru} 
& 15 
& $1.00\times$ & \multicolumn{1}{c|}{$\mathbf{26.43}\times$} 
& $1.00\times$ & \multicolumn{1}{c|}{$\mathbf{22.85}\times$} 
& $1.15$ 
\\ \\
\bottomrule
% \end{tabularx}
\end{tabular}
\end{adjustbox}
\caption{We use the White-box GA attack to produce sponge examples and measure the performance on different platforms and calculate how energy improvement factor relates to time improvement factor. The GPU readings are from NVML. GA was ran for 1000 epochs with a pool size of 1000. A detailed explanation of the results is in \Cref{sec:eval:language}.}
\label{tab:nlp-ratios}
\end{table*}

\begin{table*}[!h]
\centering
\begin{adjustbox}{scale=0.8,center}
\begin{tabular}{@{}lcccc@{}}
% \begin{tabularx}{\linewidth}{lcc *{6}{Y} }
\toprule
&  & \multicolumn{3}{c}{\textbf{ASIC Energy [mJ]}} \\

& Input size & Natural & Random & Sponge \\
\\
\midrule
\multicolumn{5}{l}{\textit{\underline{SuperGLUE Benchmark with \cite{liu2019roberta}}}} \vspace{3mm}\\
\multirow{3}{*}{CoLA} 
& 15 
& $504.93 \pm 1.07$ & $566.58 \pm 2.74$ & $583.56 \pm 0.00$
\\
& 30
& $508.73 \pm 1.87$ & $634.24 \pm 4.06$ & $669.20 \pm 0.00$
\\
& 50
& $511.43 \pm 3.64$ & $724.48 \pm 5.12$ & $780.57 \pm 0.59$
\\ \\

\multirow{3}{*}{MNLI}
& 15
& $509.19 \pm 1.45$ & $570.10 \pm 2.82$ & $586.43 \pm 0.00$
\\
& 30 
& $514.00 \pm 2.07$ & $638.78 \pm 3.89$ & $672.07 \pm 0.00$
\\
& 50 
&  $519.51 \pm 2.79$ & $728.82 \pm 5.26$ & $783.18 \pm 0.75$ 
\\ \\

\multirow{3}{*}{WSC}
& 15 
& $510.84 \pm 8.84$ & $1008.59 \pm 192.22$ & $2454.89 \pm 68.06$
\\
& 30 
& $573.78 \pm 140.12$ & $2319.05 \pm 502.31$ & $5012.75 \pm 154.24$
\\
& 50 
& $716.96 \pm 223.75$ & $5093.42 \pm 1020.34$ & $10192.41 \pm 347.32$
\\ \\
\midrule

\multicolumn{5}{l}{\textit{\underline{WMT14/16 with \cite{ott2018scaling}}}}\vspace{3mm}\\
\multirow{1}{*}{En$\rightarrow$Fr} 
& 15
& $1793.84 \pm 356.29$ & $4961.56 \pm 1320.84$ & $8494.36 \pm 166.22$\\
\\ 

\multirow{1}{*}{En$\rightarrow$De} 
& 15
& $1571.59 \pm 301.69$ & $2476.18 \pm 1586.95$ & $48446.29 \pm 0.06$ \\
\\ \\
\midrule

\multicolumn{5}{l}{\textit{\underline{WMT18 with \cite{edunov2018understanding}}}}\vspace{3mm}\\

\multirow{1}{*}{En$\rightarrow$De} 
& 15
& $1624.05 \pm 352.99$ & $2318.50 \pm 296.09$ & $49617.68 \pm 0.02$ \\
\\ \\
\midrule

\multicolumn{5}{l}{\textit{\underline{WMT19 with \cite{ng2019facebook}}}}\vspace{3mm}\\
\multirow{1}{*}{En$\rightarrow$Ru} 
& 15 
& $1897.19 \pm 607.30$ & $5380.20 \pm 2219.24$ & $47931.11 \pm 0.00$ \\
\\ \\
\bottomrule
% \end{tabularx}
\end{tabular}
\end{adjustbox}
\caption{We use the White-box GA attack to produce sponge examples and measure the consistency of ASIC results. GA was ran for 1000 epochs with a pool size of 1000. A detailed explanation of the results is in \Cref{sec:eval:language}.}
\label{tab:nlp-ratios-stds}
\end{table*}

\end{document}